\documentclass[11pt,twocolumn, clsnofoot]{IEEEtran}
\usepackage{mathrsfs}

\usepackage{algorithm}
\usepackage{algorithmic}

\usepackage{graphicx}
\usepackage{subfigure}

\ifCLASSOPTIONcompsoc
\else
\fi
\ifCLASSINFOpdf
\else
\fi
\usepackage{graphicx}
\usepackage{subfigure}
\usepackage{cite}
\newtheorem{theorem}{Theorem}

\newtheorem{assumption}{Assumption}

 \usepackage{amssymb}
\usepackage{graphicx}
\usepackage{amsmath,amssymb,amsfonts}
\usepackage{color,xcolor}
\usepackage{graphicx}
\usepackage{manfnt}
\usepackage{pgf,tikz}
\usepackage{graphicx}
\usepackage{subfigure}
\usepackage{epsfig}
\begin{document}

\title{Constructive  neural network learning}

\author{Shaobo Lin, Jinshan Zeng$^*$, and Xiaoqin Zhang 
\IEEEcompsocitemizethanks{\IEEEcompsocthanksitem S. Lin  and X.
Zhang are  with the College of Mathematics and Information Science,
Wenzhou University, Wenzhou 325035, China, and J. Zeng is with the
College of Computer Information Engineering, Jiangxi Normal
University, Nanchang, 330022, China.(Email: sblin1983@gmail.com,
jsh.zeng@gmail.com, zhangxiaoqinnan@gmail.com. $*$\textit{Corresponding author: J. Zeng.})} }

\IEEEcompsoctitleabstractindextext{%
\begin{abstract}
In this paper, we aim at developing scalable neural network-type learning systems. Motivated by the idea of ``constructive neural networks'' in approximation theory, we focus on ``constructing'' rather than ``training'' feed-forward neural networks (FNNs) for learning, and propose a novel FNNs learning system called the constructive feed-forward neural network (CFN). Theoretically, we prove that the proposed method not only overcomes the classical saturation problem for FNN approximation, but also reaches the optimal learning rate when the regression function is smooth, while the state-of-the-art learning rates established for traditional FNNs are only near optimal (up to a logarithmic factor). A series of numerical simulations are provided to show the efficiency and feasibility of CFN via comparing with the well-known regularized least squares (RLS) with Gaussian kernel and extreme learning machine
(ELM).

\end{abstract}


\begin{IEEEkeywords}
Neural networks, constructive neural network learning,
generalization error, saturation
\end{IEEEkeywords}}

\maketitle

\IEEEdisplaynotcompsoctitleabstractindextext

\IEEEpeerreviewmaketitle

\section{Introduction}\label{Sec. 1}

Technological innovations bring a profound impact on the process of
knowledge discovery. Collecting data of huge size  becomes
increasingly frequent in diverse areas of modern scientific research
\cite{Zhou2014}.
 When the amount of data is huge, many
traditional modeling strategies such as  kernel methods
\cite{Taylor2004} and neural networks \cite{Hagan1996} become
infeasible due to their heavy computational burden. Designing
effective and efficient approaches to extract useful information
from massive data has been a recent focus in   machine learning.

Scalable learning systems based on kernel methods   have been
designed for this purpose, such as the low-rank approximations of
kernel matrices  \cite{Bach2013}, incomplete Cholesky decomposition
\cite{Fine2002}, early-stopping of iterative regularization
\cite{Yao2007} and distributed learning equipped with a
divide-and-conquer scheme \cite{Zhang2014}. However,
most of the existing methods including
the gradient-based method such as the back propagation
\cite{Rumelhart1986}, second order optimization
\cite{Wilamowski2010}, greedy search \cite{Barron2008},
and
the randomization method  like random vector functional-link networks
\cite{Pao1989},   echo-state  networks \cite{Jaeger2004}, extreme
learning machines \cite{Huang20061}
fail in generating scalable
neural network-type learning systems of high quality, since the
gradient-based method  usually  suffers  from the  local minima
  and time-consuming problems, while the random method  sometimes
brings an additional ``uncertainty''   and generalization capability
degeneration phenomenon \cite{Lin2015b}. In this paper, we aim at
introducing a novel scalable feed-forward neural network (FNN)
learning system to tackle massive data.

\subsection{Motivations}
%

FNN  can be mathematically represented by
\begin{equation}\label{Neu1}
                \sum_{i=1}^nc_i\sigma(a_i\cdot x+b_i), \ x\in\mathbf
                R^d,
\end{equation}
where $\sigma:\mathbf R\rightarrow\mathbf R$ is
{ an} activation function, $a_i\in\mathbf R^{d}$, $b_i\in\mathbf R$,
and $c_i\in\mathbf R$ are the inner weight, threshold and outer
weight  of  FNN, respectively. All of $a_i$, $b_i$ and $c_i$ are
adjustable in the process of training.

From  approximation theory viewpoints,   parameters of FNN can be
either determined via training \cite{Barron1993} or constructed
based on data directly \cite{Llanas2006}.
 However, the
``construction'' idea for FNN
{did not} attract researchers' attention in the machine learning
community, although various FNNs possessing optimal approximation
property have been constructed
\cite{Anastassiou2011,Cao2008,Costarelli2013,Costarelli2013a,Lin2014b,Llanas2006,Maiorov2005}.
 The main reason is that   the constructed FNN   possesses superior learning capability   for
noise-free data only, which is usually impossible for real world
applications. \begin{figure}[H]
\begin{minipage}[b]{.45\linewidth}
\centering
\includegraphics*[scale=0.32]{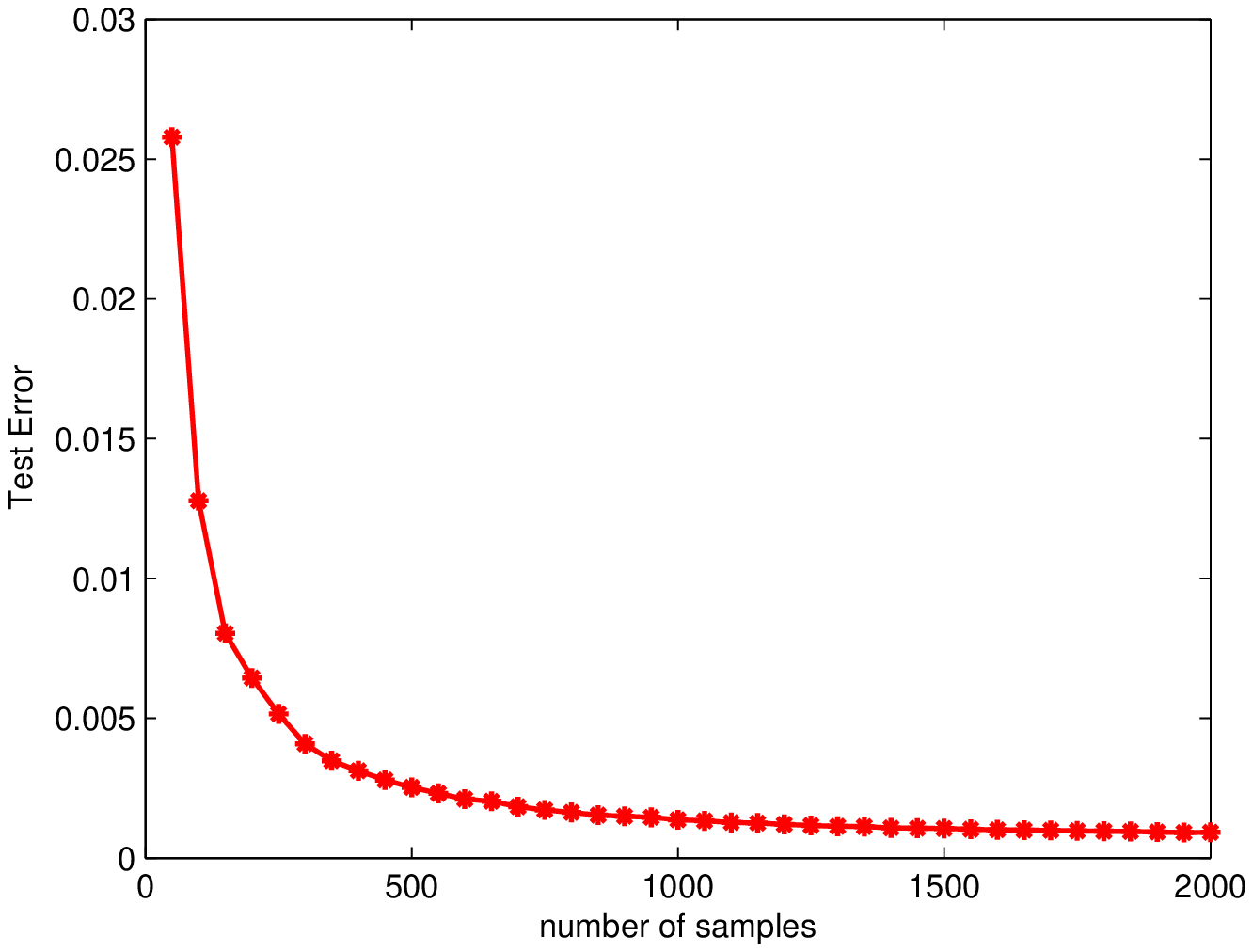}
\centerline{{\small (a) Noise-free data}}
\end{minipage}
\hfill
\begin{minipage}[b]{.45\linewidth}
\centering
\includegraphics*[scale=0.32]{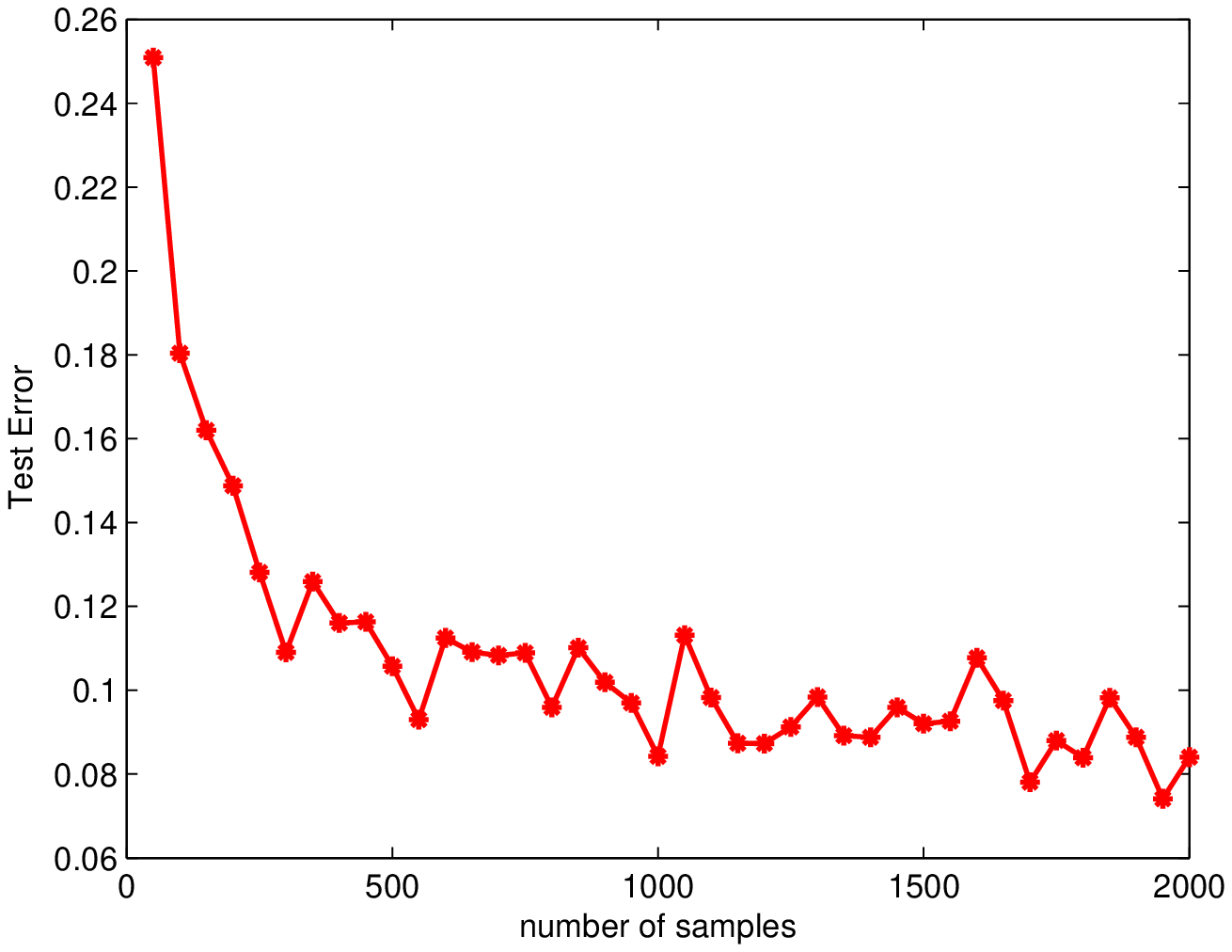}
\centerline{{\small (b) Noise data}}
\end{minipage}
\hfill \caption{Learning capability of \cite{Chen1993}'s networks
}\label{Fig:motivation}
\end{figure}

\begin{figure}[H]
\begin{minipage}[b]{.45\linewidth}
\centering
\includegraphics*[scale=0.32]{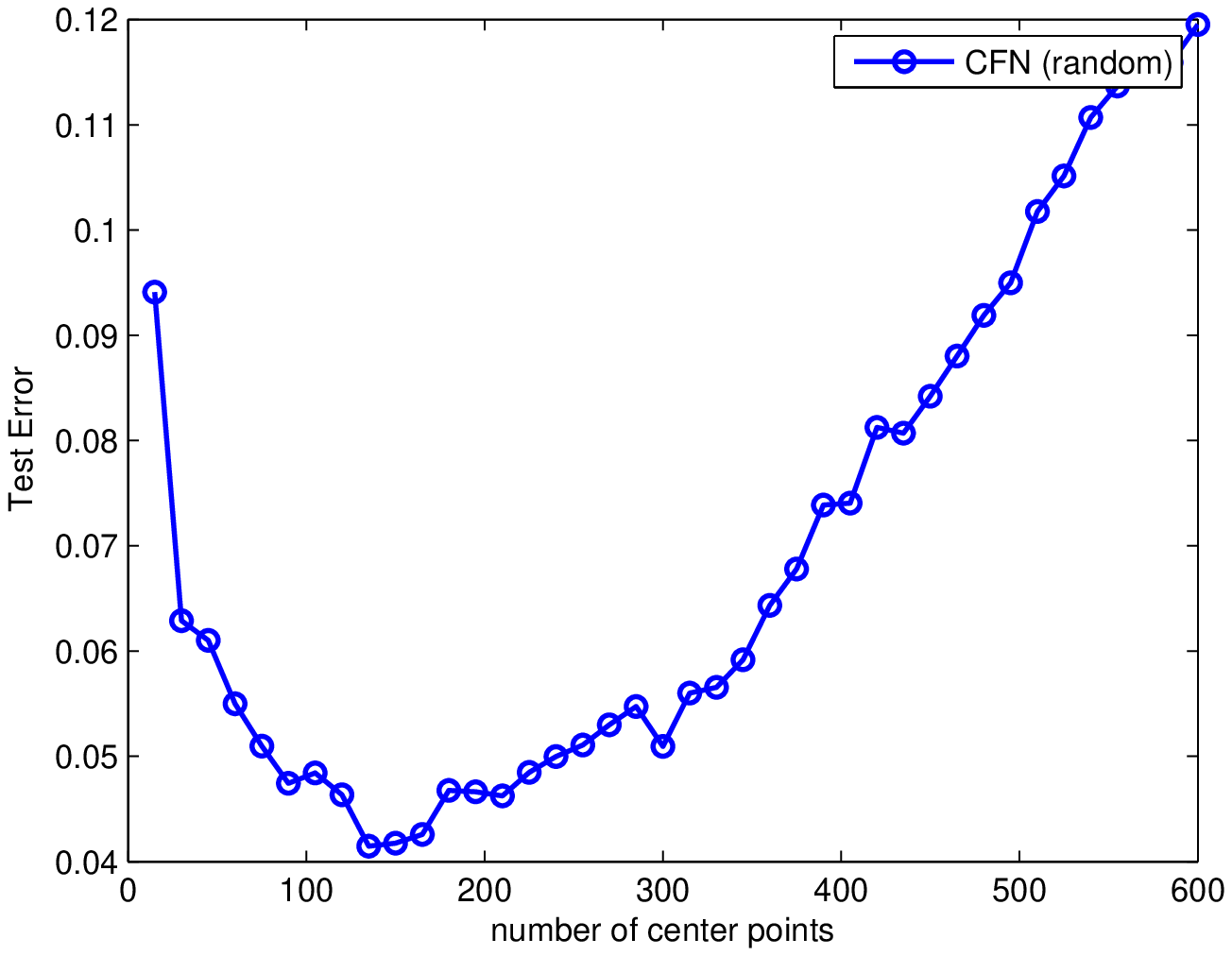}
\centerline{{\small (a) Learning by part of samples}}
\end{minipage}
\hfill
\begin{minipage}[b]{.45\linewidth}
\centering
\includegraphics*[scale=0.32]{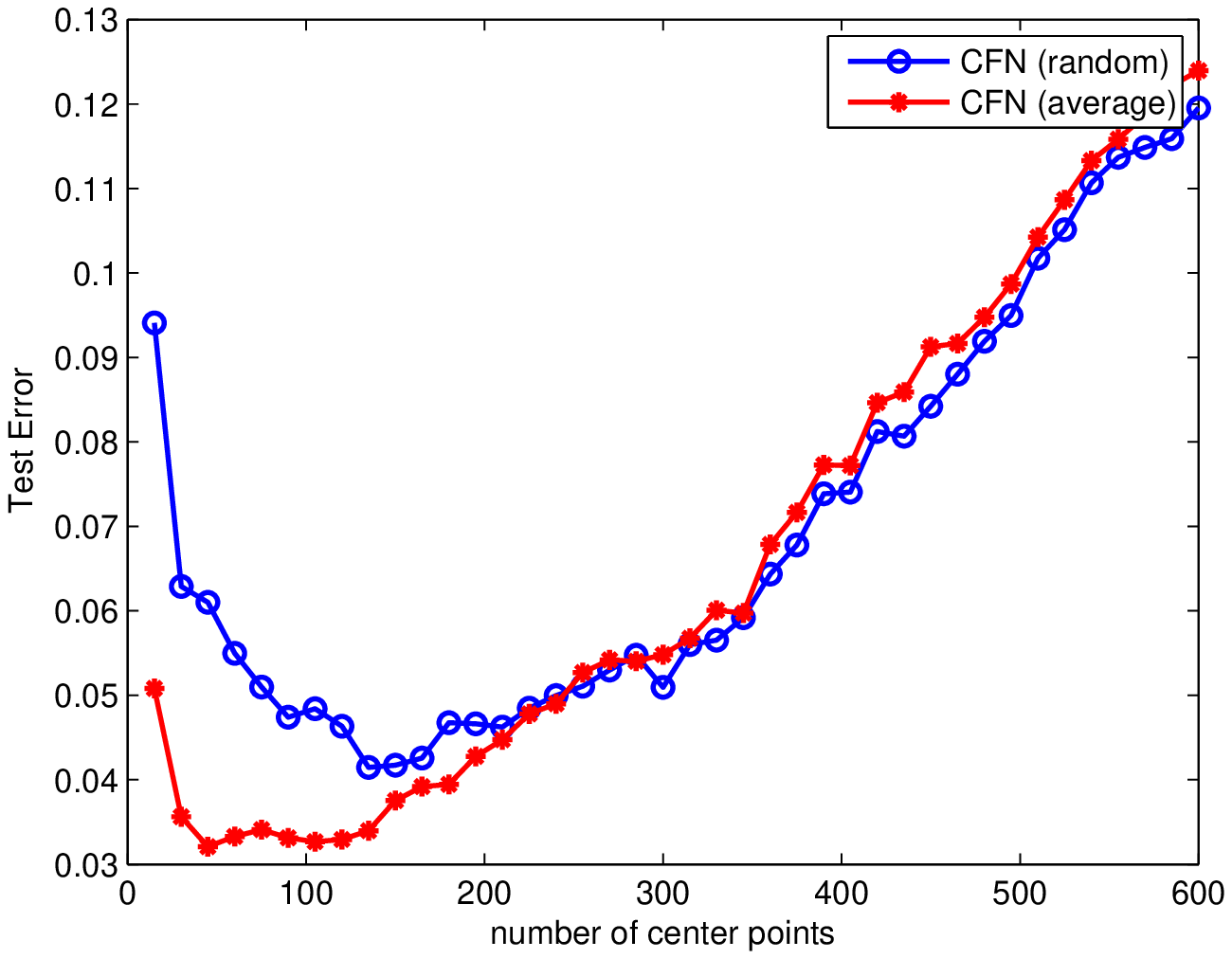}
\centerline{{\small (b) Learning through average}}
\end{minipage}
\hfill \caption{ Performance of refined construction of
\cite{Chen1993}'s networks}\label{Fig:motivation2}
\end{figure}

Fig.\ref{Fig:motivation} shows the learning performance of FNN
constructed in \cite{Chen1993}. The experimental setting in
Fig.\ref{Fig:motivation} can be found in Section \ref{Sec.4}. From
approximation to learning,  the tug of war between bias and variance
\cite{Cucker2007} dictates that besides the approximation
capability, a learning system of high quality should take  the cost
to reach the   approximation accuracy into account. Using the
constructed FNN for learning
{ is doomed } to be overfitting, since the variance is neglected in
the  literature. A preferable way to reduce the variance is to cut
down   the number of neurons. Comparing Fig.\ref{Fig:motivation2}(a)
with Fig.\ref{Fig:motivation}(b), we find that selecting part of
samples with good geometrical distribution to construct FNN
possesses better learning performance than using the whole data.

It is unreasonable   to throw away part of samples and therefore
leaves a possibility to improve the learning performance further.
Our idea stems from  the local average argument in statistics
\cite{Gyorfi2002}, showing that noise-resistent estimators can be
derived by averaging samples in a small region.   The construction
starts with selecting a set of centers (not the samples) with good
geometrical distribution and
 generating the
Voronoi partitions \cite{Eberts2015} based on them. Then,   an FNN
is constructed according to the well-developed  constructive
technique in approximation theory
\cite{Chen1993,Lin2014b,Llanas2006} by averaging outputs whose
corresponding inputs are in the same partition. As the constructed
FNN suffers from the well-known saturation phenomenon in the sense
that the learning rate cannot be improved once the smoothness of the
regression function goes beyond a specific value \cite{Gyorfi2002},
we present a Landweber-type iterative method
  to overcome   saturation
in the last step.
 As shown in
Fig.\ref{Fig:motivation2}(b), the performance of   FNN
 constructed
in such a way   outperforms the previous FNNs.

\subsection{Contributions}
 In this paper, we
  adopt
 the ideas from ``constructive neural
networks'' in approximation theory
\cite{Chen1993,Lin2014b,Llanas2006},  ``local average'' in
statistics \cite{Gyorfi2002}, ``Voronoi partition'' in numerical
analysis \cite{Wendland2005,Eberts2015} and ``Landweber-type
iteration for saturation problem'' in inverse problems
\cite{Engl1987,Engl2000} to propose a new scalable learning scheme
for FNN. In short, we aim at constructing an FNN, called the
{\underline{c}}onstructive {\underline{f}}eed-forward neural
{\underline{n}}etwork (CFN) for learning.  Our main contributions
are three folds.

Firstly, different from the previous optimization-based training
schemes \cite{Barron2008,Kohler2011}, our approach shows that
parameters of FNN can be constructed directly based on  samples,
which essentially reduces the computational burden and provides a
scalable FNN learning system to tackle massive data. The method is
novel in terms of providing a step stone from  FNN approximation to
FNN learning by using  tools in statistics, numerical analysis and
inverse problems.

The saturation problem, which was proposed in \cite{Chen1993} as an
open question, is a classical and long-standing problem of
constructive FNN approximation.  In fact,  all of FNNs constructed
in
 \cite{Anastassiou2011,Cao2008,Chen1993,Costarelli2013,Costarelli2013a,Lin2014b,Llanas2006}
suffer from the saturation problem.
 We highlight our second contribution
to   provide  an efficient iterative method to overcome the
saturation problem without  affecting  the variance very much.

Our last contribution is the feasibility verification of the
proposed CFN. We verify  both theoretical optimality and numerical
efficiency of it. Theoretically, we prove that if the regression
function is smooth, then CFN  achieves the optimal learning rate in
the sense that the upper  and lower bounds are identical and equal
to the best learning rates \cite{Gyorfi2002}.
   Experimentally, we run a
  series of numerical simulations to
verify the theoretical assertions on CFN, particularly, the good
generalization capability and low computational burden.

\subsection{Outline}

The rest of paper is organized as follows. In the next section, we
present the details of CFN.
 In Section \ref{Sec.3}, we study the theoretical behavior of CFN
 and compare it with some related work. In Section \ref{Sec.4},
some simulation results are reported to verify the theoretical
assertions. In Section \ref{sec.proof}, we prove the main results.
In
 Section \ref{Sec. Conclusion}, we
 { conclude the paper}
 and present some discussions.

\section{Construction of CFN}\label{Sec. 2}

The construction of CFN  is based on three factors: a Voronoi
partition of the input space, a partition-based distance, and
Landweber-type  iterations.

\subsection{Voronoi Partition}
Let $\mathcal X\subset\mathbf R^d$ be a compact set and
$\Xi_n:=\{\xi_j\}_{j=1}^n$ be a set of points in $\mathcal X$.
Define the mesh norm $h_{\Xi}$ and separate radius $q_{_\Xi}$
\cite{Wendland2005} of $\Xi_n$ by
$$
                 h_{\Xi}:=\max_{x\in \mathcal
                 X}\min_jd(x,\xi_j),\quad
                 q_{_\Xi}:=\frac12\min_{1\leq j\neq k\leq
                 n}d(\xi_j,\xi_k),
$$
where $d(x,x')$ denotes the Euclidean distance between $x$ and $x'$.
If there exists a constant $\tau\geq 1$ satisfying $
          {h_{\Xi}}/{q_{_\Xi}}\leq\tau,
$ then $\Xi_n$ is said to be quasi-uniform \cite{Wendland2005} with
parameter $\tau$. Throughout the paper, we assume $\Xi_n$ is a
quasi-uniform set with parameter 2\footnote{The existence of a
quasi-uniform set with parameter 2 was verified in
\cite{Narcowich2007}. Setting $\tau=2$ is only for the sake of
brevity. Our theory is feasible for arbitrary finite  $\tau$
independent of $n$, however, the numerical performance requires a
 small $\tau$.}. That is,
\begin{equation}\label{meshnorm}
    {h_\Xi}\leq 2q_{_\Xi}\leq\frac1{n^d}\leq 2h_\Xi\leq
    4q_{_\Xi}.
\end{equation}
Due to the definition of mesh norm, we get
$$
            \mathcal X\subseteq\bigcup_{j=1}^nB_j(h_\Xi),
$$
where   $B_j(r)$ denotes the Euclidean  ball  with radius $r$ and
center  ${\xi_j}$.

\begin{figure}[H]
\begin{minipage}[b]{.45\linewidth}
\centering
\includegraphics*[scale=0.32]{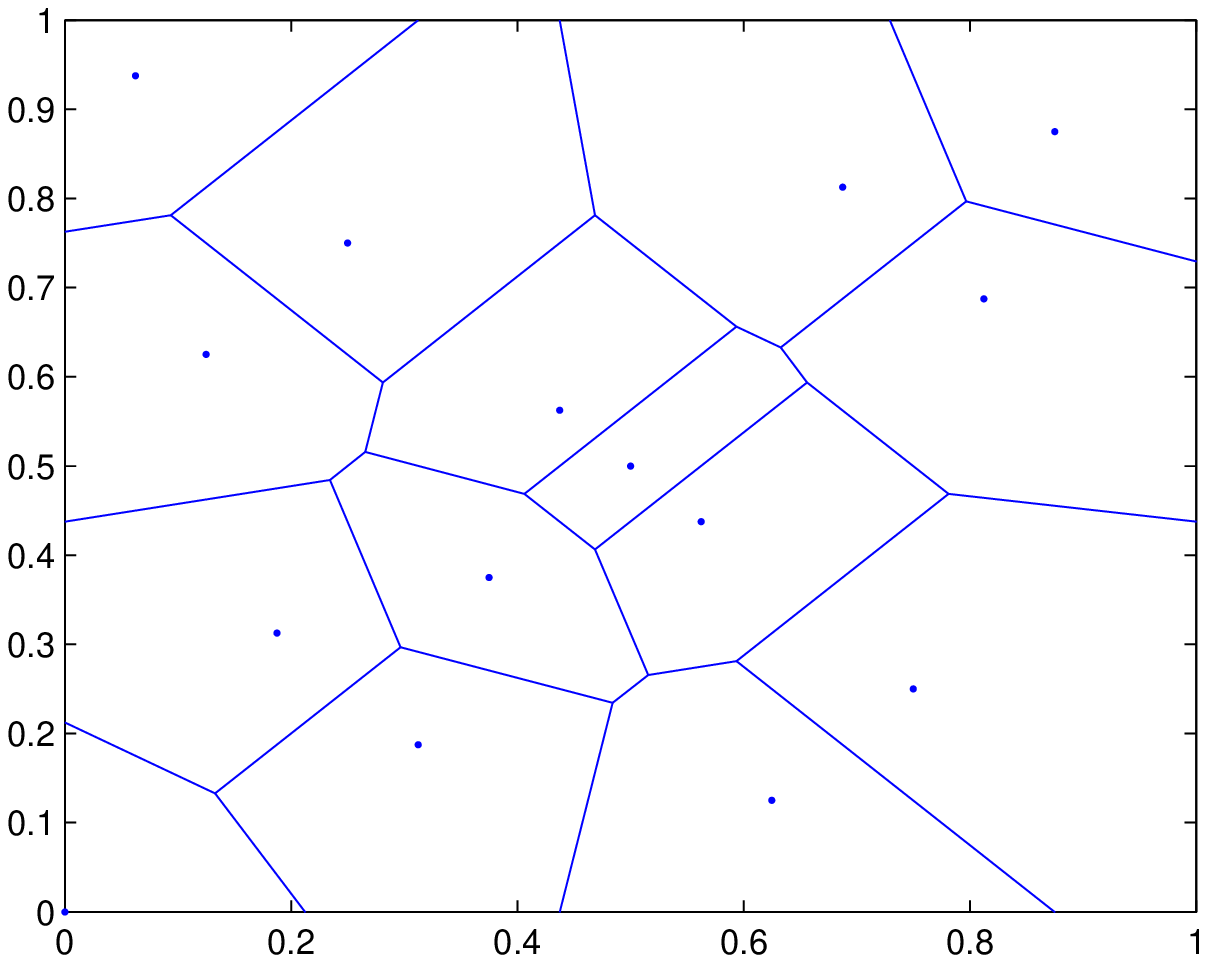}
\centerline{{\small (a) Voronoi partition}}
\end{minipage}
\hfill
\begin{minipage}[b]{.45\linewidth}
\centering
\includegraphics*[scale=0.32]{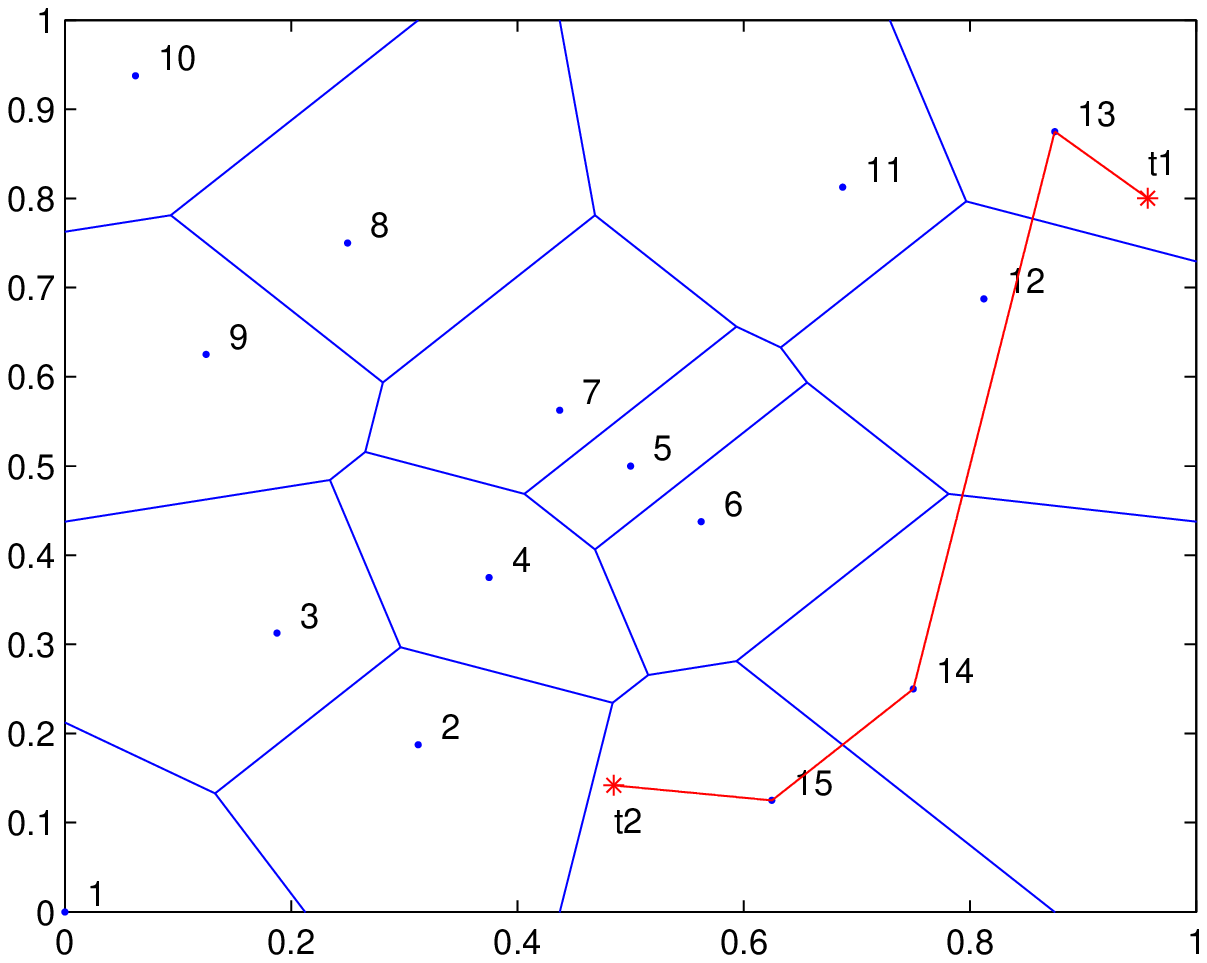}
\centerline{{\small (b) Partition-based distance}}
\end{minipage}
\hfill \caption{Voronoi partition and partition-based
distance}\label{Fig:Sobol}
\end{figure}

 A
Voronoi partition $(A_j)_{j=1}^n$ of $\mathcal X$ with respect to
$\Xi_n$ is defined (e.g. \cite{Eberts2015}) by
$$
    A_j=\left\{x\in \mathcal X:j=\min\{\arg\min_{1\leq k\leq
    n}d(x,\xi_k)\}\right\}.
$$
By definition, $A_j$ contains all $x\in \mathcal X$ such that the
center $\xi_j$ is the nearest center to $x$. Moreover, if there
exist $j_1$ and $j_2$ with $j_1<j_2$, and
$$
     d(x,\xi_{j_1})=d(x,\xi_{j_2})<d(x,\xi_k)
$$
for all $k\in\{1,2,\dots,n\}/\{j_1,j_2\}$, then $x\in A_{j_1}$,
since $j_1<j_2$.   It is obvious that $A_j\neq\varnothing$,
$A_j\subset B_j(h_{\Xi})$  for all $j\in\{1,\dots,n\}$, $A_{j_1}\cap
A_{j_2}=\varnothing$ $(j_1\neq j_2)$, and $\mathcal
X=\bigcup_{j=1}^n A_j$. Fig.\ref{Fig:Sobol} (a) presents { a
specific example} of the Voronoi partition { in $[0,1]^2$}. The
Voronoi partition is a classical partition technique for scattered
data fitting and has been widely used in numerical analysis
\cite{Wendland2005}.

\subsection{Partition-based Distance and  the first Order CFN}
To introduce the partition-based distance, we   rearrange the points
in $\Xi_n$ in the following way: let $\xi_1'$ be an arbitrary point
in $\Xi_n$, and then recursively set $\xi_{j+1}'$ satisfying
$\|\xi'_{j}-\xi'_{j+1}\|_{l^2}\leq 2h_\Xi, 1\leq j\leq n-1.$ For the
sake of brevity, we  denote  by $\Xi_n$ its rearrangement
$\{\xi_j'\}_{j=1}^n$. Let $(A_j)_{j=1}^n$ be a Voronoi partition
with respect to $\Xi_n$. Then, for
 any
$x\in\mathcal X$, there exists a unique $k_0\leq n$ such that $x\in
A_{k_0}$.
 Given any two
 points $x, x'\in X$, we define the partition-based distance
  between $x\in A_{k_0}$ and $x'\in A_{j_0}$ by
$$
      \overline{d}(x,x')=d(x,x'), \quad\mbox{if}\ k_0=j_0,
$$
and  otherwise
$$
       \overline{d}(x,x')=\sum_{j=\min\{j_0,k_0\}}^{{\max\{j_0,k_0\}}-1}
       d(\xi_j,\xi_{j+1})+d(\xi_{k_0},x)+d(\xi_{j_0},x').
$$
In the  above definition,   $\{\xi_j\}_{j=1}^n$   are used as
freight stations in computing the distance. Fig.\ref{Fig:Sobol} (b)
presents an example for the partition-based distance between $t_1$
and $t_2$ in $[0,1]^2$.

Let $S_m=(x_i,y_i)_{i=1}^m$ be the  set of samples and
$I_m=(x_i)_{i=1}^m$ be the set of inputs. Let $T_j=A_j\cap{I_m}$ be
the set of inputs  locating  in $A_j$ ($T_j$ may be  an empty set).
Denote by $T_j=\{x_1^j,\dots,x_{|T_j|}^j\}$ and its corresponding
output as $\{y_1^j,\dots,y_{|T_j|}^j\}$. The first order CFN is
given by
\begin{eqnarray}\label{sw1}
          &N^1_{n,w}({  x}):=
          \frac{\sum_{i=1}^{|T_1|}y_i^1}{|T_1|}
           +
          \sum_{j=1}^{n-1}\left(
          \frac{\sum_{i=1}^{|T_{j+1}|}y_i^{j+1}}{|T_{j+1}|}\right.\nonumber\\
          &-\left.\frac{\sum_{i=1}^{|T_{j}|}y_i^j}{|T_{j}|}\right)
          \sigma\left(w\left(\overline{d}(\xi_1,{  x})
          -\overline{d}(\xi_1,\xi_j)\right)\right),
\end{eqnarray}
where $w\in\mathbf R_+$ is a parameter  which will be determined in
the next section. Here and hereafter, we denote by $\frac{0}0=0$.

\subsection{{Iterative} Scheme for the $r$th Order CFN}

The constructed neural network in (\ref{sw1})  suffers from the
saturation problem. Indeed, it can be found in \cite{Chen1993} that
the approximation rate of (\ref{sw1})  cannot exceed $n^{-2/d}$, no
matter how smooth the regression function is. Such a saturation
phenomenon also exists for the Tikhonov regularization algorithms
\cite{Engl2000} in inverse problems and Nadaraya-Watson kernel
estimate in statistics \cite{Gyorfi2002}. It was shown in
\cite{Engl2000} and \cite{Park2009} that the saturation problem can
be avoided by iteratively learning the residual. Borrowing the ideas
from \cite{Engl2000} and \cite{Park2009}, we introduce an iterative
scheme to avoid the saturation problem of CFN.

Let $N_{n,w}^1(\cdot)$ be defined by (\ref{sw1}). For
$k=1,2,\dots,r-1$, the
 iterative
scheme is processed  as follows.

(1) Compute the residuals $e_{i,k}=y_i-N^{k}_n(x_i)$,
$i=1,2,\dots,m$.

(2) Fit $U^k_{n,w}$ to the data $\{{ x}_i,e_{i,k}\},$ $i=1,\dots,m$,
i.e.,
\begin{eqnarray}\label{sw-update}
          &U^k_{n,w}({  x}):= \frac{\sum_{i=1}^{|T_1|}e_{i,k}^1}{|T_1|}
          +
          \sum_{j=1}^{n-1}\left(
          \frac{\sum_{i=1}^{|T_{j+1}|}e_{i,k}^{j+1}}{|T_{j+1}|}\right.\nonumber\\
          &\left.-\frac{\sum_{i=1}^{|T_{j}|}e_{i,k}^j}{|T_{j}|}\right)
          \sigma\left(w\left(\overline{d}(\xi_1,{  x})
          -\overline{d}(\xi_1,\xi_j)\right)\right).
\end{eqnarray}

(3) Update $N_{n,w}^{k+1}({  x})=N_{n,w}^k({  x})+ U_{n,w}^k({ x})
$.

We obtain the $r$th order CFN $N_{n,w}^{r}$
 after repeating the above procedures $r-1$ times.

\subsection{Summary of   CFN}

\begin{algorithm}[H]\caption{CFN}\label{alg1}
\begin{algorithmic}
\STATE {{\bf Step 1 (Initialization)}: Given data
$S_m=\{(X_i,Y_i):i=1,\dots,m\}$, the  iteration number $r$, the
number of neurons $n$, the activation function $\sigma$ and the
parameter $w$.}

 \STATE{ {\bf Step 2 (Sampling)}: Select $n$ quasi-uniform points $\Xi_n:=\{\xi_j\}_{j=1}^n$
  in
 $\mathcal X$.}

 \STATE{ {\bf Step 3 (Rearranging)}: Rearrange  $n$ quasi-uniform points
 in $\Xi_n$ such that $d(\xi_j,\xi_{j+1})\leq \frac{2}{n^{1/d}}.$}

 \STATE{ {\bf Step 4 (Voronoi partition)}: Present a Voronoi partition with respect to $\Xi_n$ and
to get a set of subsets $(A_j)_{j=1}^n$ satisfying
 $A_j\cap A_k=\varnothing$, and $\cup_{j=1}^nA_j=\mathcal X$.}

 \STATE{ {\bf Step 5 (Constructing of the first order CFN)}:
 Define the first order CFN  by (\ref{sw1}).}

\STATE{ {\bf Step 6 (Initializing for iteration)}: For $k\in\mathbf
N$,
 define  $e_{i,k}=y_i-N^{k}_n(x_i)$ and generate a new set of sample
 $(x_i,e_{i,k})_{i=1}^m$.}

\STATE{ {\bf Step 7 (Iterative updating function)}:
 Define the updating function $U_{n,w}^k$ by (\ref{sw-update}).}

\STATE{ {\bf Step 8 (Updating)}:
 Define the $(k+1)$-th order CFN by
  $N_{n,w}^{k+1}({  x})=N_{n,w}^k({  x})+ U_{n,w}^k({
x}) $.}

\STATE{ {\bf Step 9 (Iterating)}: Increase $k$ by one and repeat
Step 6-Step 8 if $k< r$, otherwise,  the algorithm stops. }
\end{algorithmic}
\end{algorithm}

The proposed CFN can be formulated in
Algorithm \ref{alg1}.  In Step 2, we focus on selecting $n$
quasi-uniform points in $\mathcal X$. Theoretically, the
distribution of $\Xi_n$ significantly affects the approximation
capability of CFN, as well as the learning performance. Therefore,
$\Xi_n$ is arranged the more uniform, the better. In practice, we
use some well developed low-discrepancy sequences such as the Sobol
sequence \cite{Bratley1988} and the Halton sequence
\cite{Atanassov2004}  to generate $\Xi_n$. As there are some
existing matlab codes (such as the ``sobolset'' comment for the
Sobol sequence), we use  the Sobol sequence in this paper, which
requires $\mathcal O(n\log n)$ floating computations to generate $n$
points. In Step 3, to implement the rearrangement, we
 use the greedy
 scheme via searching the nearest point one by one\footnote{This
 greedy scheme sometimes generates a rearrangement that $d(\xi_j,\xi_{j+1})\leq
 c_0n^{-1/d}$ with $c_0>2$. We highlight that  the constant $2$ in the algorithm is
only for the sake of theoretical convenience.}. It requires
$\mathcal O(n^2)$ floating computations. In Step 4, it requires
 $\mathcal O(n^2)$ floating computations to generate a Voronoi partition.
 In Step 5,  the partition-based distance  requires
$\mathcal O(n)$ floating computations. Then it can be deduced from
(\ref{sw1}) that there are $\mathcal O(mn)$ floating computations in
this step. From Step 6 to Step 9, there is an
  iterative
process and
  it requires $\mathcal O(rmn+rn^2)$ floating computations.
Summarily,
 the total floating computations of Algorithm \ref{alg1} are of the order $\mathcal O(rmn+rn^2)$.

\section{Theoretical Behavior}\label{Sec.3}
In this section, we analyze the theoretical behavior of   CFN in the
framework of statistical learning theory \cite{Cucker2007} and
compare it with some   related work.

\subsection{Assumptions and Main Result}
Let $S_m=(x_i,y_i)_{i=1}^m\subset\mathcal Z:=\mathcal
X\times\mathcal Y$ be a set of samples drawn independently
  according to an unknown joint distribution $\rho$, which satisfies
$
                    \rho(x,y)=\rho_X(x)\rho(y|x).
$ Here, $\rho_X$ is the marginal distribution and $\rho(y|x)$ is the
conditional distribution. Without loss of generality, we  assume
$|y_i|\leq M$ almost surely for some positive number $M$.
   The performance of an estimate, $f$, is measured by the
generalization error
$$
                     \mathcal E(f):=\int_{\mathcal Z}(f(x)-y)^2d\rho,
$$
which is minimized by the regression function defined by
$$
                     f_\rho(x):=\int_{\mathcal Y}yd\rho(y|x).
$$
Let $L^2_{\rho_{_X}}$ be the Hilbert space of $\rho_X$ square
integrable functions on $\mathcal X$, with norm  $\|\cdot\|_\rho.$
{According to \cite{Cucker2007},} there holds
\begin{equation}\label{equality}
                     \mathcal E(f)-\mathcal E(f_\rho)=\|f-f_\rho\|^2_\rho.
\end{equation}

To state the main result, some assumptions on  the regression
function and the activation function should be imposed.  Let
$s=u+\beta$ for some $u\in\mathbf N$ and $0<\beta\leq 1$. A function
$f:\mathcal X\rightarrow\mathbf R$ is called $s$-smooth if for every
$\alpha=(\alpha_1,\dots,\alpha_d)$, $\alpha_l\in\mathbf N$,
$\sum_{l=1}^d\alpha_l=u$ and for all  $
       x,x'\in\mathcal X$, the partial derivative
$\frac{\partial^uf}{\partial x_1^{\alpha_1}\cdots\partial
x_d^{\alpha_d}}$ exists and satisfies
$$
    \left|\frac{\partial^uf}{\partial x_1^{\alpha_1}\cdots\partial
       x_d^{\alpha_d}}(x)-\frac{\partial^uf}{\partial x_1^{\alpha_1}\cdots\partial
       x_d^{\alpha_d}}(x')\right|\leq C[d(x,x')]^\beta
$$
for some universal  positive  constant $C$. Let $\mathcal F^{s}$ be
the set of all $s$-smooth functions.

\begin{assumption}\label{ASSUMPTION 1}
$f_\rho\in\mathcal F^{s}$ for some $s>0$.
\end{assumption}

 Assumption \ref{ASSUMPTION 1} describes the
smoothness of $f_\rho$,
 and is a regular assumption in  learning theory. It has been adopted in
\cite{Gyorfi2002,Kohler2011,Lin2015b,Liu2015,Maiorov2006a} to
quantify the learning performance of various neural network-type
learning systems.
  Let $\mathcal
M(\Theta)$ be the class of all Borel measures $\rho$ on $\mathcal Z$
such that $f_\rho\in\Theta$. Let $\mathbf G_m$ be the set of all
estimators derived from the sample $S_m$. Define
$$
          e_m(\Theta):=\inf_{f_S\in\mathbf G_m}\sup_{\rho\in \mathcal
          M(\Theta)}\mathbf E\left\{\|f_\rho-f_{S}\|^2_\rho\right\}.
$$
Obviously, $e_m(\Theta)$ quantitatively measures the quality of
$f_{S}$.  It can be found in \cite[Th.3.2]{Gyorfi2002} or
\cite[Eq.(3.26)]{Devore2006} that
\begin{equation}\label{baseline}
            e_m(\mathcal F^s)\geq C_0m^{-\frac{2s}{2s+d}},\ m=1,2,\dots,
\end{equation}
where  $C_0$ is a constant depending only on $C$, $M$, $s$  and $d$.
If an $S_m$-based estimator $f_S$  reaches the bound
$$
           \sup_{\rho\in \mathcal
          M(\mathcal F^s)}\mathbf E\left\{\|f_\rho-f_{S}\|^2_\rho\right\}\leq
          C_1m^{-\frac{2s}{2s+d}},
$$
where $C_1$ is a constant independent of $m$, then $f_m$ is
rate-optimal for $\mathcal F^s$.

Let $\sigma:\mathbf
R\rightarrow\mathbf R$ be
 a
sigmoidal function, i.e.,
$$
           \lim_{t\rightarrow +\infty }\sigma (t)=1,\quad \lim_{t\rightarrow -\infty
           }\sigma
           (t)=0.
$$
Then, there exists a  $K>0$ such that
\begin{equation}\label{definition K for sigmoidal 1}
                 |\sigma(t)-1|<n^{-(s+d)/d} \quad \mbox{if} \quad
                 t\geq K,
\end{equation}
and
\begin{equation}\label{definition K for sigmoidal 2}
                 |\sigma(t)|<n^{-(s+d)/d}
                 \quad
                 \mbox{if}
                 \quad
                 t\leq-K.
\end{equation}
For a positive number $a$, we denote by $[a]$, $\lceil a\rceil$, and
$\lfloor a\rfloor$ the integer part of $a$, the smallest integer not
smaller than $a$ and the largest integer smaller than $a$.

\begin{assumption}\label{ASSUMPTION 2}
(i)  $\sigma $ is a bounded sigmoidal function.\\
(ii) For $s>0$, $\sigma$ is at least $\lfloor s\rfloor$
differentiable.
\end{assumption}

 Conditions (i) and (ii) are
mild. Indeed, there are numerous examples satisfying (i) and (ii)
for arbitrary $s$, such as the logistic function
$$
               \sigma(t)=\frac{1}{1+e^{-t}},
$$
 hyperbolic tangent sigmoidal function
$$
         \sigma(t)=\frac12(\tanh(t)+1)
$$
with $\tanh(t)=(e^{2t}-1)/(e^{2t}+1)$,   arctan sigmoidal function
$$
         \sigma(t)=\frac1{\pi}\arctan(t)+\frac12,
$$
and  Gompertz function
$$
        \sigma(t)=e^{-a e^{-bt}}
$$
with $a,b>0$.

The following Theorem \ref{THEOREM1} is the main result of this
paper, which illustrates the rate optimality of CFN.

\begin{theorem}\label{THEOREM1}
Let $s>0$, $r\in\mathbf N$ with $r\geq s$ and $N_{n,w}^r$ be the
estimator defined by Algorithm \ref{alg1}.
 Under Assumptions \ref{ASSUMPTION 1} and \ref{ASSUMPTION 2}, if $n\sim
m^{\frac{d}{2s+d}}$ and $w\geq 4Kn^{1/{d}}$ with $K$ satisfying
(\ref{definition K for sigmoidal 1}) and (\ref{definition K for
sigmoidal 2}), then there exists a constant $C_2$ independent of $m$
or $n$ such that
\begin{equation}\label{theorem1}
        C_0m^{-\frac{2s}{2s+d}}\leq \sup_{f_\rho\in \mathcal F^{s}}\mathbf
        E\left\{\|N_{n,w}^r-f_\rho\|_\rho^2\right\}
        \leq
        C_2m^{-\frac{2s}{2s+d}}.
\end{equation}
\end{theorem}

\subsection{Remarks and Comparisons}

There are three parameters in CFN: the number of centers $n$, the
parameter $w$ and the number of iterations $r$.  Theorem
\ref{THEOREM1} shows that if some priori information of the
regression function is known, i.e., $f_\rho\in\mathcal F^s$, then
all these parameters can be determined. In particular, we can set
$w\geq4Kn^{1/d}$, $r=\lceil s \rceil$, and
$n=\left[m^{d/(2s+d)}\right].$  $K$ in (\ref{definition K for
sigmoidal 1}) and (\ref{definition K
for sigmoidal 2}) depends on $s$, $\sigma$ and $n$. So $%
K $ can be specified when $\sigma$ and $n$ are given. For example,
if the logistic function is utilized, then  $w\geq
4\frac{(s+d)n^{1/d}\log n}d$ is preferable.

However, in real world applications, it is difficult to check the
smoothness of the regression function. We provide some suggestions
about the parameter selection. Since Theorem \ref{THEOREM1} holds
for all $w\geq 4Kn^{1/d}$, $w$ can be selected to be sufficiently
large. The {iterative} scheme is imposed only for overcoming the
saturation problem and  regression functions in real world
applications are usually not very smooth. We find that a few
iterations (say, $r\leq 5$) are commonly sufficient. Thus, we   set
$r\in[1,5]$ in real world applications and use the cross-validation
\cite{Gyorfi2002} to fix it. The key parameter is $n$, which is also
crucial for almost all neural network-type learning systems, since
$n$ reflects the trade-off between bias and variance. We also  use
the cross-validation  to determine it. Compared with the
optimization-based neural network-type learning systems, the total
computational complexity of CFN (with cross-validation) is much
 lower, since the computational complexity of   CFN is
of order $\mathcal O(mn)$ for fixed $r$ and that of
optimization-based methods is at least $\mathcal O(mn^2)$
\cite{Huang20061}.

The saturation problem of FNN approximation was proposed as an open
question by Chen \cite{Chen1993}. It can be found in
\cite{Anastassiou2011,Atanassov2004,Cao2008,Costarelli2013,Costarelli2013a,Lin2014b}
 that the saturation problem for constructive neural network
approximation
  has not
been settled, since all these results were built upon the assumption
that the regression function belongs to $\mathcal F^s$ with $0<s\leq
1$.   However, for optimization-based FNN, the saturation problem
 did not
exist as shown in the prominent work
 \cite{Maiorov2000,Maiorov2005}. In the present paper,
we succeed in settling the saturation problem by using the proposed
 iterative scheme. Theorem \ref{THEOREM1}  states  that if
$f_\rho\in \mathcal F^{s}$, then CFN is rate-optimal for all $s>0$,
since
 its
learning rate can reach the base line (\ref{baseline}).

Finally, we compare Theorem \ref{THEOREM1} with two related
theoretical results about learning performance of optimization-based
FNN learning \cite{Maiorov2006a} and extreme learning machine (ELM)
\cite{Liu2015}. Denote
$$
        \mathcal N^M_n:=\left\{f=\sum_{k=1}^nc_k\phi(a_k\cdot
        x+b_k):\|f\|_\infty\leq M
        \right\},
$$
where $a_k\in\mathbf R^d$, $b_k,c_k\in\mathbf R$. Define
\begin{equation}\label{least squares}
          f_{1,S}=\arg\min_{f\in\mathcal
          N^M_n}\frac1m\sum_{i=1}^m(f(x_i)-y_i)^2.
\end{equation}
 Maiorov \cite{Maiorov2006a} proved that for some activation
function  $\phi$, if $f_\rho\in\mathcal F^s$ and $n\sim
m^{d/(s+d)}$, then there holds
\begin{equation}\label{least squares results}
        C_0m^{-\frac{2s}{2s+d}}\leq \sup_{f_\rho\in \mathcal F^{s}}\mathbf
        E\left\{\|f_{1,S}-f_\rho\|_\rho^2\right\}
        \leq
        C_3m^{-\frac{2s}{2s+d}}\log m,
\end{equation}
where $C_3$ is a constant independent of $m$ or $n$. It should be
noted from  (\ref{least squares}) that $f_{1,S}$ is obtained by solving a nonlinear least squares
problem,
 which generally requires higher computational complexity than CFN.
Furthermore, comparing
(\ref{theorem1}) with (\ref{least squares results}), we find that
CFN is rate-optimal while (\ref{least squares}) is   near
rate-optimal (up to a logarithmic factor).

Denote
$$
        \mathcal N^R_n:=\left\{f=\sum_{k=1}^nc_k\phi(a^*_k\cdot
        x+b^*_k): c_k\in\mathbf R
        \right\},
$$
where $a^*_k\in\mathbf R^d$, $b^*_k\in\mathbf R$ are randomly
selected according to a distribution $\mu$.  The ELM  estimator is
 defined by
\begin{equation}\label{elm}
          f_{2,S}=\arg\min_{f\in\mathcal
          N^R_n}\frac1m\sum_{i=1}^m(f(x_i)-y_i)^2.
\end{equation}
It is easy to see that the optimization problem (\ref{elm}) is a
linear problem and can be solved by using the pseudo-inverse
technique \cite{Huang20061}. The theoretical performance of ELM was
justified in \cite{Liu2015}, which asserts that for some activation
function,  if $f_\rho\in\mathcal F^s$ and $n\sim m^{d/(s+d)}$, then
there holds
\begin{equation}\label{elm results}
        \mathbf E_\mu  \mathbf E\left\{\|f_{2,S}-f_\rho\|_\rho^2\right\}
        \leq
        C_4m^{-\frac{2s}{2s+d}}\log m,
\end{equation}
where $C_4$ is a constant independent of $m$ or $n$. It follows from
(\ref{baseline}) and (\ref{elm results}) that (\ref{elm}) is  near
rate-optimal
in the sense of expectation, since there is an additional $\mathbf
E_\mu$ in (\ref{elm results}) due to the randomness of $f_{2,S}$.
Compared with (\ref{elm results}), our result in (\ref{theorem1})
shows that CFN can remove the logarithmic factor and avoid the
randomness of ELM.

\section{Simulations}\label{Sec.4}
In this section,  we verify our  theoretical assertions via a series
of numerical simulations. When $d=1$, the rearranging step (Step 3
in Alg.\ref{alg1}) is trivial. Thus, we divide the simulations into
two cases:  $d=1$ and $d>1$.   All numerical studies are implemented
by using MATLAB R2014a on a Windows personal computer with Core(TM)
i7-3770 3.40GHz CPUs and RAM 4.00GB. Throughout the simulations, the
logistic activation function is used and the statistics are averaged
based on 20 independent trails.

\subsection{Simulations for $d=1$}

In the  simulations, we generate training data from the following
model:
\begin{equation}\label{model1}
             Y = f(X)+ \varepsilon,
\end{equation}
where $\varepsilon$ is the   Gaussian noise with variance $0.1$ and
independent of $X$.   The centers are $n$ equally spaced points in
$[-1,1]$. The test   data are sampled from $Y=f(X)$. Let
$$
       f_1(x)= 1+\frac{80}{3}x^2 - 40 x^3 +15 x^4 + \frac{8}{3} x^5 +
       20x^2\log(|x|),
$$
and
$$
     f_2(x)=(1-x)_+^5(8x^2+5x+1),
$$
where $a_+=a$ when $a\geq 0$ and $a_+=0$ when $a<0$. It is easy to
check that  $f_1\in\mathcal F^{1}$ but $f_1\notin\mathcal F^2$ and
$f_2\in\mathcal F^4$ but $f_2\notin \mathcal F^5$.

\begin{figure}[H]
\begin{minipage}[a]{.45\linewidth}
\centering
\includegraphics*[scale=0.32]{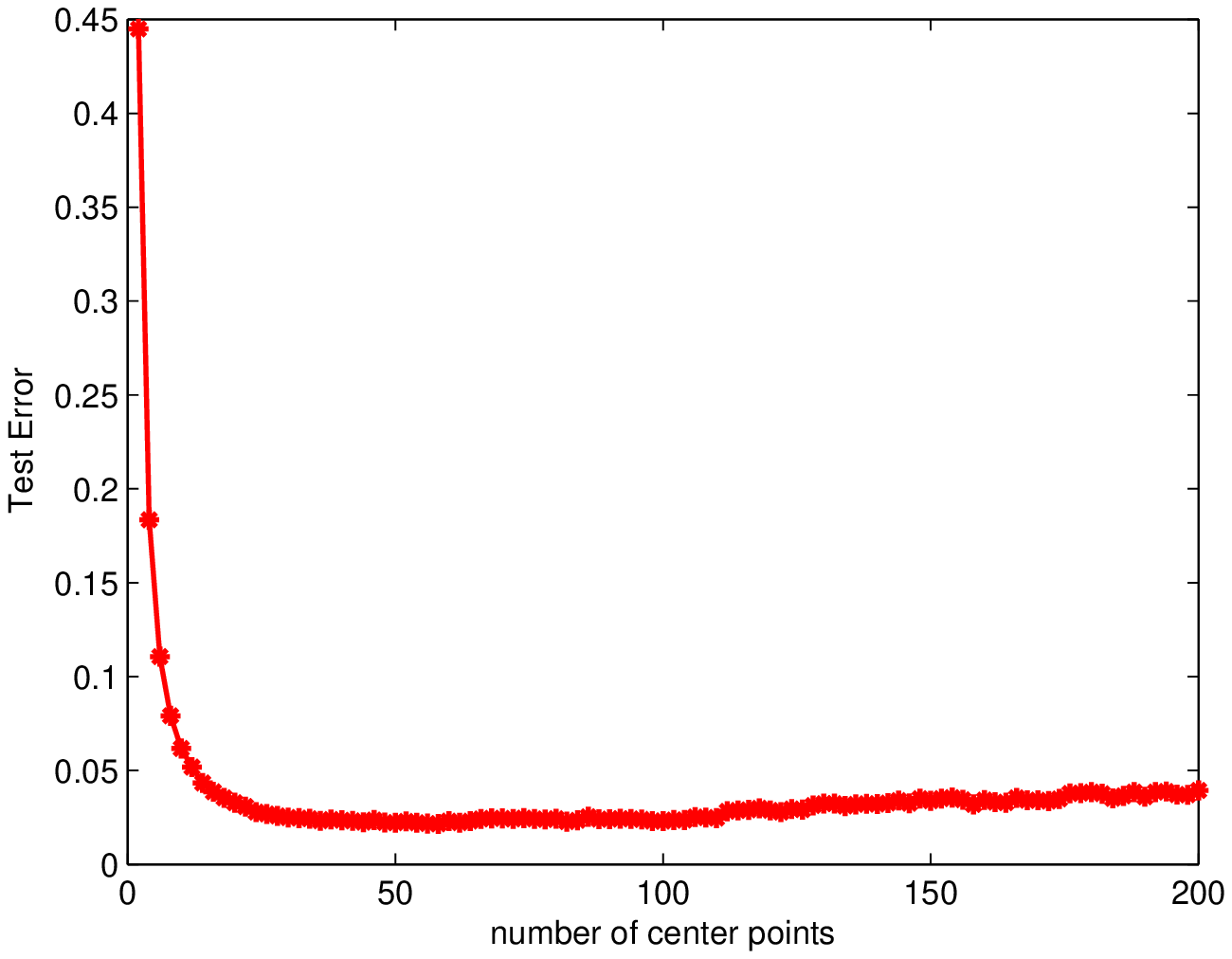}
\centerline{{\small (a) Role of $n$ for $f_1$}}
\end{minipage}
\hfill
\begin{minipage}[a]{.45\linewidth}
\centering
\includegraphics*[scale=0.32]{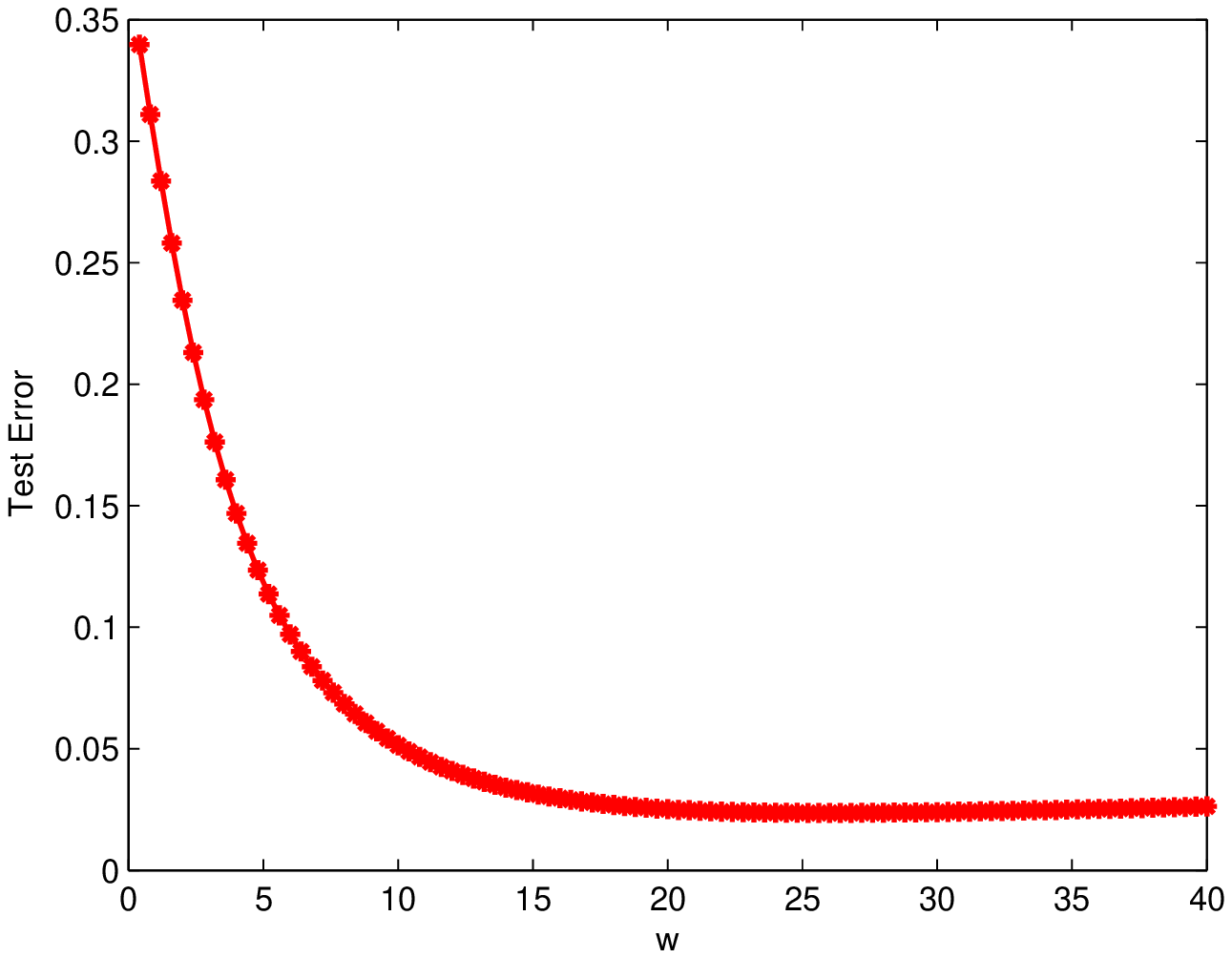}
\centerline{{\small (b) Role of $w$ for $f_1$}}
\end{minipage}
\begin{minipage}[a]{.45\linewidth}
\centering
\includegraphics*[scale=0.32]{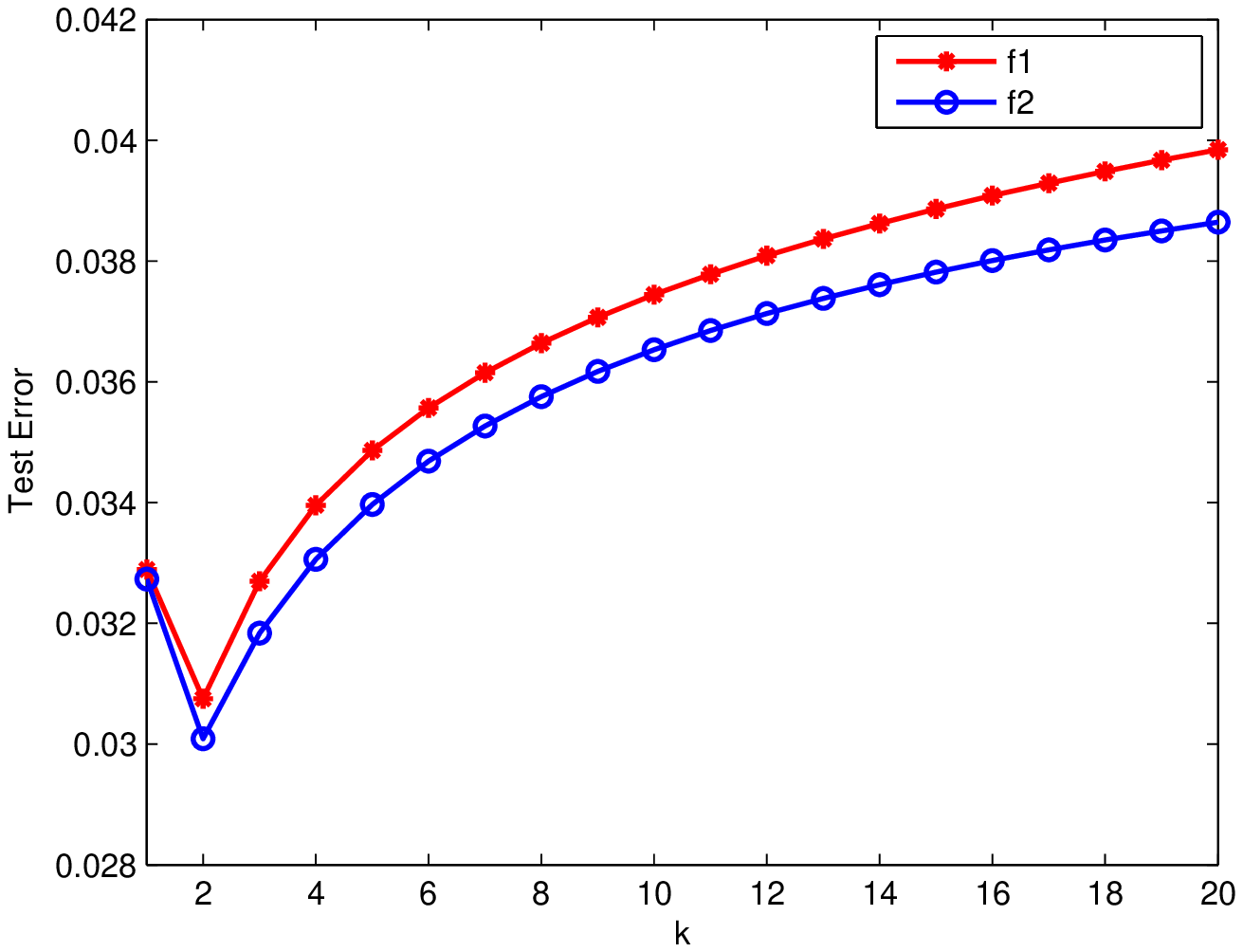}
\centerline{{\small (c) Role of $r$ for $f_1$ and $f_2$}}
\end{minipage}
\hfill
\begin{minipage}[a]{.45\linewidth}
\centering
\includegraphics*[scale=0.32]{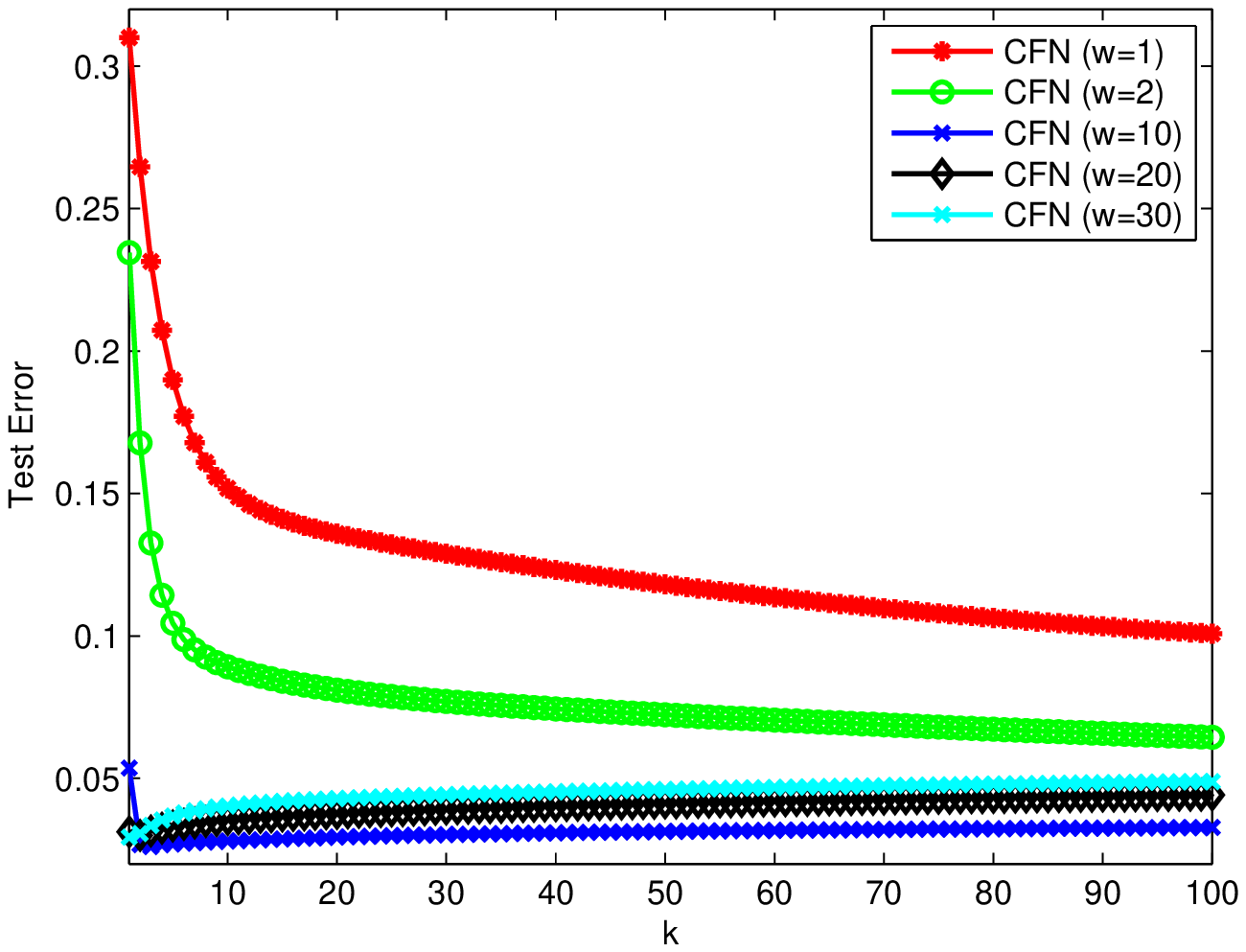}
\centerline{{\small (d) Small  $w$ and large $r$ for $f_1$}}
\end{minipage}
\hfill \caption{Roles of parameters}\label{Fig:one dimensional 1}
\end{figure}

 In the first
simulation, we   numerically study the roles of three parameters
$n$, $w$, and $r$. Since we are interested in the role of a
specified parameter,  the other two parameters   are selected to be
optimal according to the test data directly.  In this simulation,
 the number of training and test samples are 1024 and 1000.
The results of the simulation are shown in Fig.\ref{Fig:one
dimensional 1}.
 Fig.\ref{Fig:one dimensional 1} (a) describes the relation between the test
error and the  number of centers. From Fig.\ref{Fig:one dimensional
1} (a), it follows that there exists an optimal $n$   ranged in
$[10,20]$ minimizing the test error, which coincides with the
theoretical assertions $n\sim m^{d/(2s+d)}=1024^{1/3}$.
Fig.\ref{Fig:one dimensional 1} (b) demonstrates  the relation
between the test error and $w$. It can be found in Fig.\ref{Fig:one
dimensional 1} (b) that after a crucial value of $w$, the test error
does not vary with respect to $w$, which coincides { with the}
assertion in Theorem \ref{THEOREM1}. Fig.\ref{Fig:one dimensional 1}
(c) presents the relation between the test error and the number of
iterations. It is shown in Fig.\ref{Fig:one dimensional 1} (c) that
such a scheme is feasible and can improve the learning performance.
Furthermore, there is  an optimal $r$ in the range of $[1,5]$
minimizing the test error (the average value of $r$ is $2.2$ via
average for 20 trails). This also verifies our assertions in
Sec.\ref{Sec.3}. Fig.\ref{Fig:one dimensional 1} (d) presents
another application of the iterative scheme in CFN. It is shown in
Fig.\ref{Fig:one dimensional 1} (d) that once the parameter $w$ is
selected to be small, then we can use large  $r$ to reduce the test
error.

\begin{table}[H]
\addtolength{\tabcolsep}{-4pt}
\begin{center}
 \caption{Comparisons of learning performance for $d=1$.}\label{Table: t1}
\begin{tabular}{|c|c|c|c|c|}\hline
          & TrainRMSE & TestRMSE& TrainingTime& TestTime    \\ \hline
  \multicolumn{5}{|c|}{$f_1$} \\ \hline
  CFN        & 0.3305(0.0070)   &{\bf 0.0182(0.0035)} & 0.51[0.01] &0.91      \\ \hline
  RLS     & 0.3303(0.0070)   & 0.0203(0.0064) & 9.95[0.19] &8.22   \\ \hline
  ELM  & {\bf 0.3302(0.0070)}   & 0.0223(0.0072) &{\bf 0.42[0.01]} &{\bf 0.31}   \\ \hline
 \multicolumn{5}{|c|}{$f_2$}               \\ \hline
  CFN       &  0.3304(0.0070)   &{\bf 0.0199(0.0037)} &0.48[0.01] &0.88   \\ \hline
  RLS     &0.3303(0.0070)   &  0.0204(0.0067) &10.6[0.21] &9.14       \\ \hline
  ELM   & {\bf 0.3302(0.0070)}   &0.0225(0.0072) &{\bf 0.42[0.01]} &{\bf 0.33}     \\ \hline
 \end{tabular}
 \end{center}
 \end{table}

In the second simulation,  we  compare CFN with two widely used
learning schemes. One is the regularized least square
(RLS) with Gaussian kernel \cite{Eberts2011}, which is recognized as
the benchmark for regression. The other is  the extreme learning
machine (ELM) \cite{Huang20061}, which is one of the most popular
neural network-type learning systems. The corresponding parameters
including the width of the Gaussian kernel and the regularization
parameters in RLS, the number of hidden neurons in ELM, and the
number of centers, the number of iterations in CFN are  selected via
five-fold cross validation. We record the mean test RMSE (rooted
square mean error)  and the mean training RMSE as TestRMSE and
TrainRMSE, respectively. We also record their standard deviations in
the parentheses. Furthermore, we record the average total training
time    as TrainTime. We also record the time of training for fixed
parameters in the bracket.  Since the training time of ELM is
different for different number of neurons, we record in  the bracket
the training time for ELM with optimal parameter. We finally record
the mean test time as TestTime. The results are reported in Table
\ref{Table: t1}. It is shown in Table \ref{Table: t1} that the
learning performance of CFN is at least not worse than RLS and ELM.
In fact, the test error of CFN is comparable with RLS but better
than ELM and the training price (training time and memory
requirement) of CFN is comparable with ELM but
{lower} than RLS. This verifies the feasibility of  CFN and
coincides with our theoretical assertions.

\subsection{Simulations for $d>1$}

When $d>1$, the strategies of selecting
  centers and rearrangement operator are required. In this part,
we draw the Sobol sequence to build up the set of centers and then
use a greedy
 strategy to rearrange them. Specifically, we start
 with  an arbitrary point in the Sobol sequence,
then select the next point to be the nearest point of the current
point, and then repeat this procedure until all the points are
rearranged. We
 show in the following Fig.\ref{Fig:high dimensional1} that such a
greedy  scheme
  essentially cuts down the maximum distance between arbitrary  two adjacent
  points (max dist). For comparison, we also exhibit the change of
  the minimum distance (min dist).
\begin{figure}[H]
\begin{minipage}[b]{.45\linewidth}
\centering
\includegraphics*[scale=0.32]{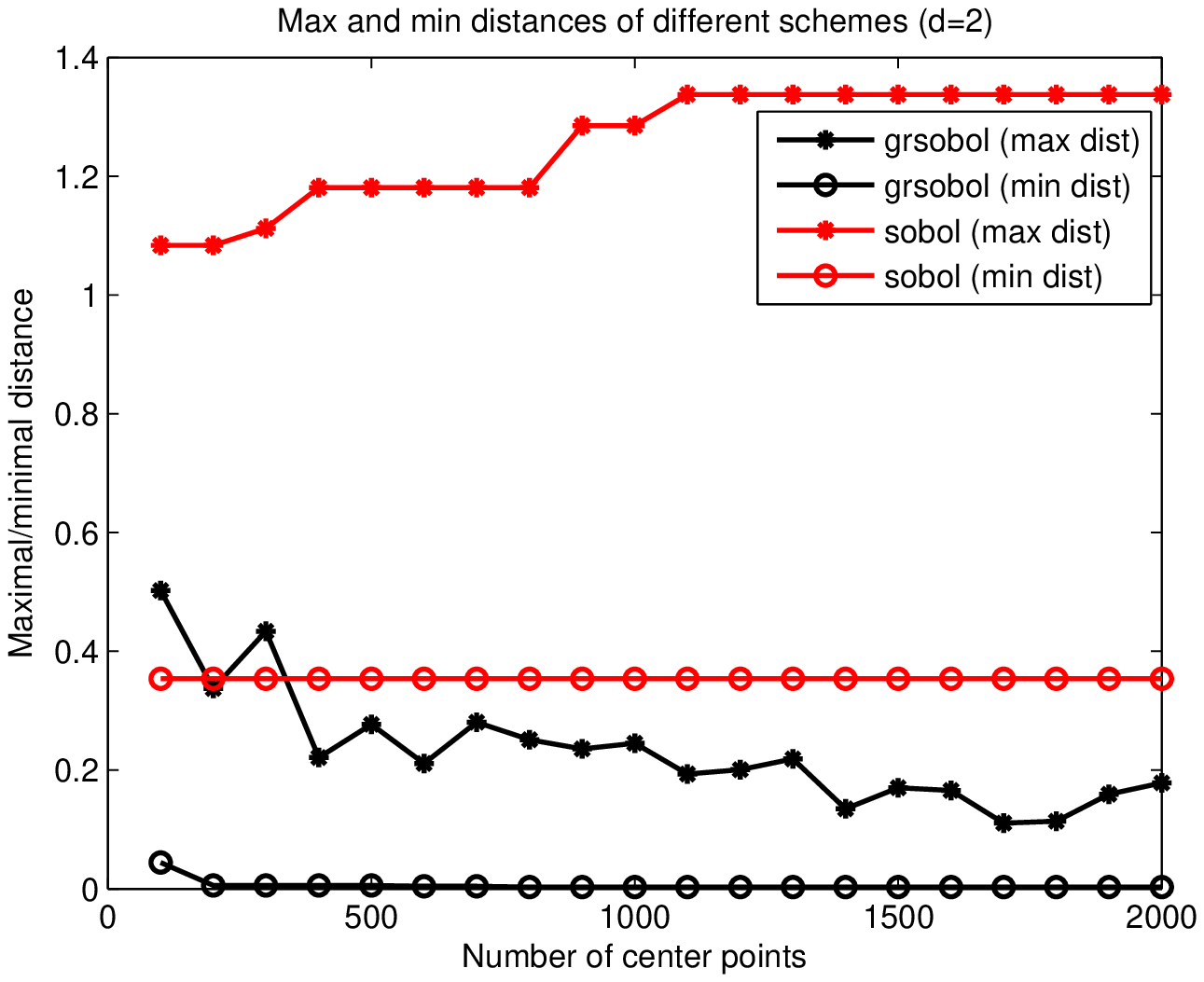}
\centerline{{\small (a) $d=2$}}
\end{minipage}
\hfill
\begin{minipage}[b]{.45\linewidth}
\centering
\includegraphics*[scale=0.32]{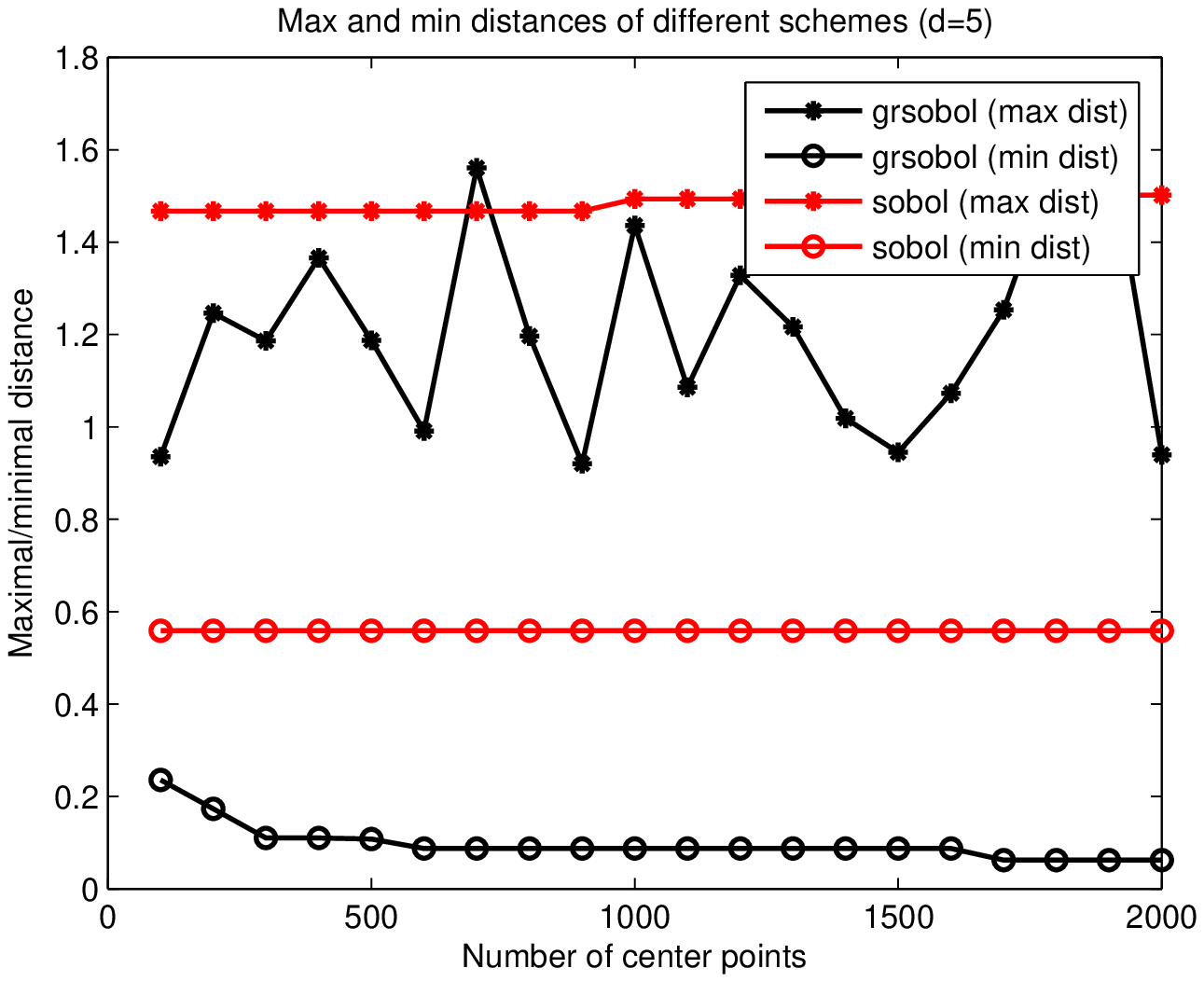}
\centerline{{\small (b) $d=5$}}
\end{minipage}
\hfill \caption{Distribution of centers}
 \label{Fig:high dimensional1}
\end{figure}

We demonstrate the feasibility of CFN by comparing it with RLS and
ELM for various regression functions. Let
\begin{align*}
          f_3(x) = (1-\|x\|_2)_{+}^6(35/3\|x\|_2^2 + 6\|x\|_2 +1),
\end{align*}
\begin{align*}
        f_4(x)=
        &(1-\|x\|_2)_{+}^7(35\|x\|_2^6 + 245 \|x\|_2^5 + 720 \|x\|_2^4\nonumber\\ & + 1120 \|x\|_2^3
        + 928 \|x\|_2^2 + 336 \|x\|_2 + 48),
\end{align*}
and
\begin{align*}
f_5(x) = & 1+\frac{80}{3} \|x\|_2^5 - 40 \|x\|_2^3 + 15 \|x\|_2^4 + \frac{8}{3}\|x\|_2^5 \nonumber\\
&+ 20 \|x\|_2^2 log(\|x\|_2),
\end{align*}
where $\|x\|_2$ denotes the Euclidean norm of the vector $x$.
 It is easy to check that for $d=2,3,5$, $f_5 \in\mathcal F^1$ but
 $f_5\notin\mathcal F^2$ and for $d=2,3$,  $f_3 \in\mathcal F^4$ but
 $f_3 \notin\mathcal F^5$ and for $d=5$, $f_4 \in\mathcal F^4$ but
 $f_4 \notin\mathcal F^5$.
  The simulation  results are reported
 in Table \ref{Tabel:2}. It can be found in  Table \ref{Tabel:2}
 that, similar as the one-dimensional simulations,
the learning performance of CFN is  a bit better than RLS (in time)
and ELM (in test error). This also verifies  our theoretical
assertions.

\begin{table}[htbp]
\addtolength{\tabcolsep}{-4pt}
\begin{center}
 \caption{Comparisons of learning performance for $d>1$.}\label{Tabel:2}
\begin{tabular}{|c|c|c|c|c|c|}\hline
          & TrainRMSE & TestRMSE& TrainingTime& TestTime   \\ \hline
  \multicolumn{5}{|c|}{  $f_3$, $d=2$} \\ \hline
  CFN     & 0.3303(0.010)   &{\bf 0.0341 (0.0094)} &4.67 [0.067] &1.05    \\ \hline
  RLS    & {\bf 0.3259(0.0102)}   & 0.0412(0.0047) & 10.6[0.217] & 9.43    \\ \hline
  ELM  & 0.3291(0.010)   &0.0391 (0.0055) &{\bf 0.42 [0.003]} &{\bf 0.34}  \\ \hline
 \multicolumn{5}{|c|}{$f_3$,  $d=3$ }               \\ \hline
  CFN      & 0.3312 (0.008)   &{\bf 0.0495 (0.0057)} & 4.77[0.09] &1.05        \\ \hline
  RLS  &  {\bf 0.3204(0.007)}   & 0.0593(0.0065) &11.0 [0.27] & 9.65        \\ \hline
  ELM  & 0.3253(0.007)   &0.0542 (0.0066) &{\bf 0.61 [0.01]} &{\bf 0.31}   \\ \hline
 \multicolumn{5}{|c|}{ $f_4$,  $d=5$ } \\ \hline
  CFN        & 0.3302(0.009)   & 0.0278(0.004) &4.96 [0.09] & 1.12        \\ \hline
  RLS    &  {\bf 0.3294(0.009)}   &{\bf 0.0272 (0.006)} &10.8 [0.20] & 9.79   \\ \hline
  ELM    &  0.3303(0.009)   & 0.0311(0.004) & {\bf 0.52[0.01]} &{\bf 0.35}     \\ \hline
 \multicolumn{5}{|c|}{ $f_5$,  $d=2$  } \\ \hline
  CFN      & 0.3318(0.010)   & {\bf 0.0286(0.007)} &4.73 [0.08] &0.81    \\ \hline
  RLS    & {\bf 0.3282(0.011)}   &0.0404 (0.006) &10.8 [0.22] &9.70    \\ \hline
  ELM     & 0.3302 (0.011)   &0.0359 (0.007) &{\bf 0.49 [0.01]} &{\bf 0.45}  \\ \hline
 \multicolumn{5}{|c|}{  $f_5$, $d=3$  }               \\ \hline
  CFN   &  0.3361(0.009)   &{\bf 0.0465 (0.009)} & 4.86[0.08] & 0.82      \\ \hline
  RLS   & 0.3260(0.007)   &0.0573 (0.006) &10.6 [0.21] &9.63      \\ \hline
  ELM & {\bf 0.3315(0.008)}   &0.0494 (0.004) &{\bf 0.56 [0.01]} &{\bf 0.38}   \\ \hline
 \multicolumn{5}{|c|}{   $f_5$, $d=5$} \\ \hline
  CFN   & 0.3310(0.007)   & {\bf 0.0299 (0.006)} &4.91 [0.08] & 0.80     \\ \hline
  RLS     & {\bf 0.3270(0.007)}   & 0.0400 (0.003) &10.7 [0.20] & 9.77        \\ \hline
  ELM   & 0.3329( 0.007)   & 0.0438(0.004) &{\bf 0.51 [0.01]} &{\bf 0.47}    \\ \hline
 \end{tabular}
 \end{center}
 \end{table}

\subsection{Challenge for massive data}
Since our main motivation for introducing CFN is to tackle massive
data,  we pursue the advantage of CFN for massive data sets. For
this purpose,  we set the number of training data to be
$50000,100000, 150000$ and the number of test data to be $100000,
200000, 300000$. As the memory requirements and training time of RLS
are huge, we only compare CFN with ELM in this simulation. The
simulation results are reported in Table \ref{Table:3}. It is shown
that when applied to the massive data,  the performance of  CFN is
at least not worse than   that of  ELM. In particular, CFN possesses
a slight smaller    test errors and    dominates in the training
time. The reason is that for  massive data, a large number of
neurons in ELM and CFN are required. When $r\leq 5$ and $n\leq m$,
 the computational complexities of CFN and ELM are $\mathcal
O(mn)$  and $\mathcal O(mn^2)$, respectively. Large $n$ inevitably
leads to much more training time for ELM. Thus, besides the perfect
theoretical behaviors, CFN is also better than ELM when tackling
massive data.

\begin{table}[htbp]
\addtolength{\tabcolsep}{-4pt}
\begin{center}
 \caption{Comparisons of learning performance for massive data.}\label{Table:3}
\begin{tabular}{|c|c|c|c|c|}\hline
          & TrainRMSE & TestRMSE& TrainTime& TestTime   \\ \hline

 \multicolumn{5}{|c|}{$f_1$, $d=1, m = 5\times 10^4$} \\ \hline
  CFN    & {\bf 0.3298(0.001)}   & {\bf 0.0033(5.0e-004)}  & {\bf 13.6[0.34]}    & 21.8     \\ \hline
  ELM    & {\bf 0.3298(0.001)}   & 0.0035(7.0e-004)   & 78.9[0.16] & {\bf 19.2}     \\ \hline
 \multicolumn{5}{|c|}{$f_1$, $d=1, m = 10^5$} \\ \hline
  CFN   & {\bf 0.3299(0.001)}   & 0.0029(7.3e-004) & {\bf 28.3[0.71]}   & 45.3     \\ \hline
  ELM   & {\bf 0.3299(0.001)}  & {\bf 0.0027(5.3e-004)} & 167[ 3.90] & {\bf 38.3}     \\ \hline
 \multicolumn{5}{|c|}{$f_1$, $d=1, m = 1.5\times 10^5$} \\ \hline
  CFN   & {\bf 0.3303(0.001)}   & {\bf 0.0021(4.8e-004)} & {\bf 42.5[1.27]}  & 67.8    \\ \hline
  ELM   & {\bf 0.3303(0.001)}   & 0.0022(6.6e-004) & 252[7.65] & {\bf 55.8}     \\ \hline
  \multicolumn{5}{|c|}{$f_5$, $d=5, m = 5\times 10^4$} \\ \hline
  CFN    & 0.3309(0.001)   & {\bf 0.0161(0.005)} & {\bf 95.8[4.51]}  & 44.6     \\ \hline
  ELM    & {\bf 0.3304(0.001)}   & 0.019(0.001) & 134[2.70] & {\bf 36.5}    \\ \hline
 \multicolumn{5}{|c|}{$f_5$, $d=5, m = 10^5$} \\ \hline
  CFN   & 0.3302(0.001)   & {\bf 0.0131(5.0e-04 )}   & {\bf 191[8.93]} & 89.1     \\ \hline
  ELM   & {\bf 0.3299(0.001)}   & 0.0159(7.0e-04 ) & 260[7.05] & {\bf 71.6}     \\ \hline
 \multicolumn{5}{|c|}{$f_5$, $d=5, m = 1.5\times 10^5$} \\ \hline
  CFN   & 0.3307(0.001)   & {\bf 0.0122(3.9e-04)}   & {\bf 291[13.2]} & 134     \\ \hline
  ELM   & {\bf 0.3305(0.001)}   & 0.0138(6.6e-04) & 399[9.81]   & {\bf 108}    \\ \hline
 \end{tabular}
 \end{center}
 \end{table}

\section{Proof of Theorem \ref{THEOREM1}}\label{sec.proof}
  The lower bound of (\ref{theorem1})
can be derived directly by (\ref{baseline}). Therefore, it suffices
to prove the upper bound of (\ref{theorem1}). For $1\leq k\leq r$,
define
$$
      \widetilde{N_{n,w}^k}(x):=\mathbf
       E\left[N_{n,w}^k(x)|X_1,\dots,X_m|\right].
$$
Then,
\begin{eqnarray}\label{Error decomposition}
        &&\mathbf
        E\left[\|N_{n,w}^r-f_\rho\|_\rho^2\right] \\
        &=&
        \mathbf E\left[\| N_{n,w}^r-\widetilde{N_{n,w}^r}\|_\rho^2\right]
        +
        \mathbf E\left[\|\widetilde{N_{n,w}^r}-f_\rho\|_\rho^2\right],\nonumber
\end{eqnarray}
since $\mathbf E\langle N_{n,w}^r-\widetilde{N_{n,w}^r},\widetilde{N_{n,w}^r}-f_\rho\rangle_\rho=0.$
 We call the first term in the righthand of (\ref{Error decomposition})
 as the sample error (or estimate error) and the second
term as the approximation error.

\subsection{Bounding Sample Error}
We divide the bounding of sample error into two cases:   $r=1$ and
$r>1$. We first
  consider
the case $r=1$. Since $f_\rho(x)=\mathbf
E[Y|X=x]$, we have
\begin{eqnarray}\label{expentatial CNNL}
          & \widetilde{N_{n,w}^1}(x)
          = \frac{\sum_{i=1}^{|T_1|}f_\rho(X_i^1)}{|T_1|}
          +
          \sum_{j=1}^{n-1}\left(\frac{\sum_{i=1}^{|T_{j+1}|}
          f_\rho(X_i^{j+1})}{|T_{j+1}|}\right.\nonumber\\
          &-
          \left.\frac{\sum_{i=1}^{|T_{j}|}f_\rho(X_i^j)}{|T_{j}|}\right)
          \sigma\left(w\left(\overline{d}(\xi_1,{  x})
          -\overline{d}(\xi_1,\xi_j)\right)\right).
\end{eqnarray}
Denote
$$
          g_j:=\frac{\sum_{i=1}^{|T_{j}|}Y_i^j}{|T_{j}|}
          =\sum_{i=1}^m Y_i\frac{I_{\{X_i\in A_j\}}}{\sum_{l=1}^m I_{\{X_l\in A_l\}}},
$$
and
$$
          g_j^*:=\frac{\sum_{i=1}^{|T_{j}|}f_\rho(X_i^j)}{|T_{j}|}
          =\sum_{i=1}^m f_\rho(X_i)
          \frac{I_{\{X_i\in A_j\}}}{\sum_{l=1}^m I_{\{X_l\in
          A_l\}}},
$$
 where $I_A$ denotes the indicator function of set $A$.
Define further
$$
                     c_1(x):=1-\sigma(w\overline{d}(\xi_1,x)),
$$
$$
                     c_n(x):=\sigma\left(w(\overline{d}(\xi_1,x)
                     -\overline{d}(\xi_1,\xi_{n-1})\right),
$$
and
\begin{eqnarray*}
                    &c_j(x)
                    :=\sigma\left(w(\overline{d}(\xi_1,x)
                    -\overline{d}(\xi_1,\xi_{j-1}))\right)\\
                    &-
                    \sigma\left(w(\overline{d}(\xi_1,x)
                    -\overline{d}(\xi_1,\xi_{j}))\right),
                      \ 2\leq j\leq n-1,
\end{eqnarray*}
We have
\begin{equation}\label{operator1}
                 N_{n,w}^1 (x)=\sum_{j=1}^ng_jc_j(x),
\end{equation}
and
\begin{equation}\label{operator2}
                \widetilde{N_{n,w}^1}(x)=\sum_{j=1}^ng_j^*c_j(x).
\end{equation}
Since $A_{j_1}\cap A_{j_2}=\varnothing$ ($j_1\neq j_2$) and
$\mathcal X=\bigcup_{j=1}^n A_j$,   for arbitrary $x$, there is a
unique $k_0$ such that $x\in A_{k_0}$. Without loss of generality,
we assume $k_0\geq 2$.  Then it follows from (\ref{meshnorm}) and
the definition of $\overline{d}(\xi_1,{ x})$ that
$$
                  \overline{d}(\xi_1,{ x})-
                  \overline{d}({\xi_1,\xi_k})
                  \geq \frac14n^{-1/d}, \ 1\leq k\leq k_0-1,
$$
$$
                 |\overline{d}(\xi_1,{  x})
                 -\overline{d}(\xi_1,\xi_{k_0})|\leq 2n^{-1/d},
$$
and
$$
                  \overline{d}(\xi_1,{x})-\overline{d}({\xi_1,\xi_k})
                  \leq -\frac14n^{-1/d}, \ k\geq k_0+1.
$$
Due to (\ref{definition K for sigmoidal 1}), (\ref{definition K for
sigmoidal 2})  and $w\geq 4Kn^{1/d}$, we have when $k\leq k_0-1$
\begin{equation}\label{1}
                  \left|\sigma\left(w\left(\overline{d}(\xi_1,{ x})
                  -\overline{d}(\xi_1,\xi_k)\right)\right)-1\right|<n^{-(s+d)/d},
\end{equation}
and when  $k\geq
                  k_0+1$,
\begin{equation}\label{2}
                  \left|\sigma\left(w\left(\overline{d}(\xi_1,{x})
                  -\overline{d}(\xi_1,\xi_k)\right)\right)\right|<n^{-(s+d)/d}.
\end{equation}
Then    for  arbitrary $j\leq k_0-1$  and $j\geq k_0+2$, (\ref{1}),
(\ref{2}) and the definition of $c_j(x)$ yield
$$
         |c_j(x)|
          \leq 2n^{-(s+d)/d}.
$$
Since $|Y_j|\leq M$ almost surely, we have $|g_j|\leq M$ and
$|g^*_j|\leq M$ almost surely $j=1,\dots,n$. As
$$
        \mathbf
        E\left[g_j|X_1,\dots,X_m\right]=g_j^*,
$$
we get
\begin{eqnarray*}
        && \mathbf
        E\left[(N_{n,w}^1(x)-\widetilde{N_{n,w}^1}(x))^2|X_1,\dots,X_m\right]\\
        &\leq&
        2M^2n^{-2s/d}\\
        &+&
        \mathbf
        E\left[
        ((g_{k_0}-g_{k_0}^*)c_{k_0}(x))^2|X_1,\dots,X_m\right]\\
        &+&
         \mathbf
        E\left[
        ((g_{k_0+1}-g_{k_0+1}^*)c_{k_0+1}(x))^2|X_1,\dots,X_m\right].
\end{eqnarray*}
We hence only need to bound
$$
         \mathbf E\left[
        ((g_{k_0}-g_{k_0}^*)c_{k_0}(x))^2|X_1,\dots,X_m\right],
$$
since
$$
\mathbf
        E\left[
        ((g_{k_0+1}-g_{k_0+1}^*)c_{k_0+1}(x))^2|X_1,\dots,X_m\right]
$$
can be bounded by using the same method.
 As $\sigma$ is bounded, we have
\begin{eqnarray*}
       &\mathbf E\left[
        ((g_{k_0}-g_{k_0}^*)c_{k_0}(x))^2|X_1,\dots,X_m\right]\\
        &\leq 4\|\sigma\|^2_\infty\mathbf E\left[
         (g_{k_0}-g_{k_0}^*)^2|X_1,\dots,X_m\right].
\end{eqnarray*}
Due to the definition of $g_{k_0}$ and $g^*_{k_0}$, we get
\begin{eqnarray*}
     &&
     \mathbf E\left[
        ((g_{k_0}-g_{k_0}^*))^2|X_1,\dots,X_m\right]\\
        &=&
         \frac{\sum_{i=1}^m(Y_i-f_\rho(X_i))^2I_{\{X_i\in
        A_{k_0}\}}}{(m\mu_m(A_{k_0}))^2}\\
         &\leq&
        \frac{4M^2}{m\mu_m(A_{k_0})}I_{\{m\mu_m(A_{k_0})>0\} },
\end{eqnarray*}
where $\mu_m(A_{k_0})$ denotes the empirical measure of $A_{k_0}$.
Then it can be found in \cite[P65-P66]{Gyorfi2002} that
$$
        \mathbf E\left[\frac{I_{\{m\mu_m(A_{k_0})>0\}}}{m\mu_m(A_{k_0})}
        \right]\leq\frac{4n}m.
$$
This implies
\begin{eqnarray*}
          &&
          \mathbf
        E\left[\|N_{n,w}^1-\widetilde{N_{n,w}^1}\|_\rho^2\right]\\
        &=&
          \mathbf E\left[\mathbf
        E\left[(N_{n,w}^1(X)-\widetilde{N_{n,w}^1}(X))^2|X,X_1,\dots,X_m\right]\right]\\
        &\leq&
        2M^2n^{-2s/d}+32\|\sigma\|_\infty^2M^2{n}/{m}.
\end{eqnarray*}
Since $n\sim m^{d/(2s+d)}$, we obtain
\begin{equation}\label{Sample error r=1}
       \mathbf
        E\left[\|N_{n,w}^1-\widetilde{N_{n,w}^1}\|_\rho^2\right]
        \leq
        C'm^{-2s/(2s+d)},
\end{equation}
where $C'$ is a constant depending only on $M$.

We then bound the sample error for $r>1$. It follows from the
definition of $ N_{n,w}^{k+1} (x)$ that
\begin{eqnarray*}
        &&N_{n,w}^{k+1} (x)
         =
        N_{n,w}^{k} (x)+ N_{n,w}^{1} (x)\\
        &-&
        \frac{\sum_{i=1}^{|T_1|}(N^{k}_{n,w}(X_i^1))}{|T_1|}
            -
          \sum_{j=1}^{n-1}\left( \frac{\sum_{i=1}^{|T_{j+1}|}
          (N^{k}_{n,w}(X_i^{j+1}))}{|T_{j+1}|}\right.\\
          &-&
          \left.
          \frac{\sum_{i=1}^{|T_{j}|}(N^{k}_{n,w}(X_i^{j}))}{|T_{j}|}\right)
          \sigma\left(w\left(\overline{d}(\xi_1,{  x})
          -\overline{d}(\xi_1,\xi_j)\right)\right),
\end{eqnarray*}
where we denotes the element of $T_j$ as
$X_1^j,X_2^j,\dots,X_{|T_j|}^j$. Since
\begin{eqnarray*}
        &&\frac{\sum_{i=1}^{|T_1|}(N^{k}_n(X_i^1))}{|T_1|}
            +
          \sum_{j=1}^{n-1}\left( \frac{\sum_{i=1}^{|T_{j+1}|}
          (N^{k}_n(X_i^{j+1}))}{|T_{j+1}|}\right.\\
          &-&
          \left.\frac{\sum_{i=1}^{|T_{j}|}(N^{k}_n(X_i^{j}))}{|T_{j}|}\right)
          \sigma\left(w\left(\overline{d}(\xi_1,{  x})
          -\overline{d}(\xi_1,\xi_j)\right)\right).
\end{eqnarray*}
is independent of $Y_1,\dots,Y_m$, and
$$
         \widetilde{N_{n,w}^k}(x)=\mathbf
         E\left[N_{n,w}^k(x)|X_1,\dots,X_m|\right],
$$
 we have
\begin{eqnarray*}
       &&\mathbf
       E\left[(N_{n,w}^{k+1}(x)-\widetilde{N_{n,w}^{k+1}}(x))^2|X_1,\dots,X_m\right]\\
       &\leq&
       2\mathbf E\left[(N_{n,w}^{1}(x)-\widetilde{N_{n,w}^{1}}(x))^2
       |X_1,\dots,X_m\right]\\
       &+&
       2\mathbf E
       \left[(N_{n,w}^{k}(x)-\widetilde{N_{n,w}^{k}}(x))^2|X_1,\dots,X_m\right].
\end{eqnarray*}
Therefore, we obtain
\begin{eqnarray*}
       \mathbf
       E\left[\|N_{n,w}^{r}-\widetilde{N_{n,w}^{r}}\|_\rho^2\right]
       \leq
       2r\mathbf E\left[\|N_{n,w}^{1}-\widetilde{N_{n,w}^{1}}\|_\rho^2
      \right].
\end{eqnarray*}
Then it follows from (\ref{Sample error r=1}) that
\begin{equation}\label{sample error r>1}
       \mathbf
        E\left[\|N_{n,w}^r-\widetilde{N_{n,w}^r}\|_\rho^2\right]
        \leq
        2rC'm^{-2s/(2s+d)}.
\end{equation}

\subsection{Bounding Approximation Error}
It is obvious that
\begin{eqnarray*}
           &&(\widetilde{N_{n,w}^r}(x)-f_\rho(x))^2\\
           &=&
           (\widetilde{N_{n,w}^r}(x)-f_\rho(x))^2I_{\{\mu_m(A_{k_0})>0\}}\\
           &+&
           (\widetilde{N_{n,w}^r}(x)-f_\rho(x))^2I_{\{\mu_m(A_{k_0})=0\}}.
\end{eqnarray*}
Due to the definition, we obtain
$$
            \| \widetilde{N_{n,w}^r}\|_\infty \leq (2\|\sigma\|_\infty+1)rM
$$
almost surely. Then it follows from \cite[P66-P67]{Gyorfi2002} that
\begin{eqnarray*}
          &&\mathbf
          E\left[(\widetilde{N_{n,w}^r}(X)-f_\rho(X))^2I_{\{\mu_m(A_{k_0})=0\}}
          \right]\\
          &\leq&
            3(2\|\sigma\|_\infty+1)^2r^2M^2\frac{n}m.
\end{eqnarray*}
Furthermore,
\begin{eqnarray*}
       &&(\widetilde{N_{n,w}^r}(x)-f_\rho(x))^2I_{\{\mu_m(A_{k_0})>0\}}\\
       &=&(\widetilde{N_{n,w}^r}(x)-f_\rho(x))^2I_{\{\mu_m(A_{k_0})>0,
       \mu_m(A_{k_0+1})>0\}}\\
       &+&
       (\widetilde{N_{n,w}^r}(x)-f_\rho(x))^2I_{\{\mu_m(A_{k_0})>0,
       \mu_m(A_{k_0+1})=0\}}.
\end{eqnarray*}
Similar method in \cite[P66-P67]{Gyorfi2002} yields
\begin{eqnarray*}
          &&\mathbf
          E\left[(\widetilde{N_{n,w}^r}(X)-f_\rho(X))^2I_{\{\mu_m(A_{k_0})>0,
          \mu_m(A_{k_0+1})=0\}}
          \right]\\
          &\leq&  3(2\|\sigma\|_\infty+1)^2r^2M^2\frac{n}m.
\end{eqnarray*}
Hence, we have
\begin{eqnarray}\label{app1}
           &&\mathbf
           E\left[(\widetilde{N_{n,w}^r}(X)-f_\rho(X))^2\right]\nonumber\\
           &\leq&
           6(2\|\sigma\|_\infty+1)^2r^2M^2\frac{n}m\\
           &+&
           \mathbf
          E\left[(\widetilde{N_{n,w}^r}(X)-f_\rho(X))^2I_{\{\mu_m(A_{k_0})>0,
          \mu_m(A_{k_0+1})>0\}}\right].\nonumber
\end{eqnarray}
To bound
$$
           \mathbf
          E\left[(\widetilde{N_{n,w}^r}(X)-f_\rho(X))^2I_{\{\mu_m(A_{k_0})>0,
          \mu_m(A_{k_0+1})>0\}}\right],
$$
we need to introduce a series of auxiliary functions. Let $X_j^*$ be
arbitrary sample in $A_j$ and $Y_j^*$ be its corresponding output.
If there
 is no point in $A_j$,  then we denote by $Y_j^*=0$, and $g(X_j^*)=0$
for arbitrary function $g$. Define
\begin{eqnarray*}
          &&L^1_{n,w}({  x})= f_\rho(X_1^*)
          +
          \sum_{j=1}^{n-1}\left( f_\rho(X_{j+1}^*)-f_\rho(X_j^*)
          \right)\\
          &\times&
          \sigma\left(w\left(\overline{d}(\xi_1,{  x})
          -\overline{d}(\xi_1,\xi_j)\right)\right),
\end{eqnarray*}
and
$$
             L^{k+1}_{n,w}({  x})= L^k_{n,w}({  x})+ V^k_{n,w}({
             x})
$$
with
\begin{eqnarray*}
          &&V^k_{n,w}({  x})
           =
            f_\rho(X^*_1)-L^{k}_n(X_1^*)\\
           &+&
          \sum_{j=1}^{n-1}\left(
          (f_\rho(X_{j+1}^*)-L^{k}_n(X_{j+1}^*))\right.\\
          &-&
          \left. (f_\rho(X^*_{j})-L^{k}_n(X^*_{j}))\right)
          \sigma\left(w\left(\overline{d}(\xi_1,{  x})
          -\overline{d}(\xi_1,\xi_j)\right)\right).
\end{eqnarray*}
Then for arbitrary  $x\in \mathcal X$  and $1\leq k\leq r$, there
holds
\begin{equation}\label{comparison}
    \left|\widetilde{N_{n,w}^k}(x)-f_\rho(x)\right|
    \leq
    \max_{X_1^*,\dots,X_n^*}\left|L^{k}_{n,w}(x)-f_\rho(x)\right|,
\end{equation}
where the maximum runs over all the possible choices of  $X_j^*\in
A_j$, $j=1,2,\dots,n$.

Due to the definition of $c_j(x)$, we get
$$
              L^{1}_{n,w}(x)=\sum_{j=1}^{n}f_\rho(X_j^*)c_j(x),\quad\mbox{
and}\
        \sum_{j=1}^nc_j(x)=1.
$$
We then prove that for arbitrary $k\geq 1$, there exists a set of
functions $\{c_j^k(x)\}_{j=1}^n$ such that
\begin{equation}\label{simple iteration}
        L^{k}_{n,w}(x)=\sum_{j=1}^{n}f_\rho(X_j^*)c^k_j(x),\
        \mbox{and}\ \sum_{j=1}^nc_j^k(x)=1.
\end{equation}
We prove (\ref{simple iteration}) by induction. (\ref{simple
iteration})   holds for $k=1$ with $c_j^1(x)=c_j(x)$. Assume
(\ref{simple iteration}) holds for $l\geq 1$, that is,
$$
        L^{l}_{n,w}(x)=\sum_{j=1}^{n}f_\rho(X_j^*)c^l_j(x),\
        \mbox{and}\ \sum_{j=1}^nc_j^l(x)=1.
$$
Obviously,
$$
           V^l_{n,w}({
           x})=\sum_{j=1}^n(f_\rho(X_j^*)-L^{l}_{n,w}(X_j^*))c_j^1(x).
$$
Therefore,
\begin{eqnarray*}
     &&L^{l+1}_{n,w}(x)
      =
     L^{l}_{n,w}(x)+ V^l_{n,w}({x})
      =
     \sum_{j=1}^{n}f_\rho(X_j^*)c^l_j(x)\\
     &+&
     \sum_{j=1}^n\left(f_\rho(X_j^*)-
     \sum_{i=1}^{n}f_\rho(X_i^*)c^l_i(X_j^*)\right)c_j^1(x)\\
     &=&
     \sum_{j=1}^nf_\rho(X_j^*)\left(c^l_j(x)+c_j^1(x)-\sum_{i=1}^nc_j^l(X_i^*)c_i^1(x)
     \right).
\end{eqnarray*}
Define
\begin{equation}\label{proof approx 1}
             c^{l+1}_j(x):=c^l_j(x)+c_j^1(x)-\sum_{i=1}^nc_j^l(X_i^*)c_i^1(x).
\end{equation}
Then it follows from
  $\sum_{j=1}^nc_j^1(x)=1$ and $\sum_{j=1}^n c_j^l(x)=1$ that
\begin{eqnarray*}
       \sum_{j=1}^n c_j^{l+1}(x)
     =2-\sum_{i=1}^n\sum_{j=1}^nc_j^l(X_i^*)c_i^1(x)=1.
\end{eqnarray*}
This proves (\ref{simple iteration}).  (\ref{simple iteration})
together with (\ref{proof approx 1}) implies
\begin{eqnarray*}
        &&L^{k+1}_{n,w}(x)-f_\rho(x)
         =
        \sum_{j=1}^nc^{k+1}_j(x)(f_\rho(X_j^*)-f_\rho(x))\\
        &=&
        \sum_{j=1}^nc^k_j(x)(f_\rho(X_j^*)-f_\rho(x))\\
        &-&
        \sum_{i=1}^nc_i^1(x)\left[\sum_{j=1}^nc^k_j(X_i^*)(f_\rho(X_j^*)
        -f_\rho(X_i^*))\right]\\
        &=&
        L^{k}_{n,w}(x)-f_\rho(x)-\sum_{j=1}^nc^1_j(x)(L^{k}_{n,w}(X_j^*)-f_\rho(X_j^*)).
\end{eqnarray*}
If we denote by $\lambda^k(x):=L^{k}_{n,w}(x)-f_\rho(x)$ with
$\lambda^0(x):=f_\rho(x)$, then
\begin{eqnarray*}
     &&\lambda^{k+1}(x)
      =
     \lambda^k(x)-\sum_{j=1}^nc^1_j(x)\lambda^k(X_j^*)\\
     &=&
     \lambda^k(x)-\lambda^k(X^*_{k_0})\\
     &-&
     \left(\lambda^k(X^*_{k_0+1})-\lambda^k(X_{k_0}^*)
     \right)
     \sigma
    \left(w(\overline{d}(\xi_1,{
     x})-\overline{d}(\xi_1,\xi_{k_0})\right)\\
     &-&\sum_{j=1}^{k_0-1}\left(\lambda^k(X^*_{j+1})-\lambda^k(X^*_{j})\right)\\
     &\times&\left(\sigma
    \left(w(\overline{d}(\xi_1,{
     x})-\overline{d}(\xi_1,\xi_j)\right)-1\right)\\
     &-&
     \sum_{j=k_0+1}^{n-1}\left(\lambda^k(X^*_{j+1})-\lambda^k(X^*_{j})\right)\\
     &\times&
     \sigma
    \left(w(\overline{d}(\xi_1,{
     x})-\overline{d}(\xi_1,\xi_j))\right).
\end{eqnarray*}
Due to (\ref{1}) and (\ref{2}), we obtain almost surely that for
arbitrary $x\in\mathcal X$
\begin{eqnarray}\label{Importance for app}
               && \lambda^{k+1}(x) \nonumber\\
               &=&
                \lambda^k(x)-\lambda^k(X^*_{k_0})
        -  (\lambda^k(X^*_{k_0+1})-\lambda^k(X_{k_0}^*)
      )\nonumber\\
      &\times&
      \sigma
    \left(w(\overline{d}(\xi_1,{
     x})-\overline{d}(\xi_1,\xi_{k_0}))\right) \pm \mathcal O(n^{-s/d}).
\end{eqnarray}
The above inequality yields
\begin{equation}\label{proof of approx  2}
         \|\lambda^{k}\|_\rho\leq
          \tilde{C}(k-\lceil
         s\rceil)n^{-s/d}+(2+2\|\sigma\|_\infty)\|\lambda^{\lceil
         s\rceil}\|_\rho,
\end{equation}
where $\tilde{C}$ is a constant independent of $n$ or $m$. Noting
$\lambda^0=f_\rho$,
 and
$f_\rho\in\mathcal F^s$, we have from (\ref{Importance for app})
that for $0\leq s\leq 1$,
$$
     \|\lambda^{\lceil s\rceil}\|_\infty\leq
     \left(C(1+\|\sigma\|_\infty)+\tilde{C}\right)n^{-s/d}.
$$
The above inequality together with   (\ref{app1}),
(\ref{comparison}), (\ref{proof of approx  2}) and $n\sim
m^{d/(2s+d)}$ yields that when $s\leq 1$, there holds
\begin{equation}\label{Approximation error s=1}
    \mathbf
    E\left[\|\widetilde{N_{n,w}^r}-f_\rho\|_\rho^2\right]\leq
    C''m^{-2s/(2s+d)},
\end{equation}
where $C''$ is a constant depending only on $C$, $\tilde{C}$, $s$
and $r$.
%
%
%
When $s> 1$,
without loss of generality, we assume $x\neq X_{k_0}^*$. We then
prove by induction that there exist continuous functions
$\alpha^{s}(x)$ and $\beta^{s}(x)$ such that for arbitrary $x\in
\mathcal X$, there holds almost surely
\begin{eqnarray}\label{proof approx 3}
          &&\lambda^{\lceil s\rceil}(x)=h_\Xi^{\lceil s\rceil}\alpha^{s}(x) \pm \mathcal
     O(n^{-s/d})\nonumber\\
          &+&
          h_\Xi^{\lceil s\rceil}\beta^{s}(x)\sigma
    \left(w(\overline{d}(\xi_1,{
     x})-\overline{d}(\xi_1,\xi_{k_0}))\right).
\end{eqnarray}
 Indeed, (\ref{Importance
for app}) together with the multivariate  mean value theorem implies
that there are $\lfloor s\rfloor-1$ times differential functions
$\alpha^1(\cdot)$ and $\beta^1(\cdot)$  such that
\begin{eqnarray*}
          &&\lambda^{\lceil s\rceil}(x)=h_\Xi\alpha^{1}(x) \pm \mathcal
     O(n^{-s/d})\nonumber\\
          &+&
          h_\Xi\beta^{1}(x)\sigma
    \left(w(\overline{d}(\xi_1,{
     x})-\overline{d}(\xi_1,\xi_{k_0}))\right).
\end{eqnarray*}
  If we assume that for all $1\leq k\leq\lceil s\rceil-1$,
there exist $\lfloor s\rfloor-k$ times differential functions
$\alpha^k(\cdot)$ and $\beta^k(\cdot)$ such that
\begin{eqnarray*}
          &&\lambda^{k}(x)=h^k\alpha^k(x)\pm \mathcal O(n^{-s/d})\\
          &+&h^k\beta^k(x)\sigma
    \left(w(\overline{d}(\xi_1,{
     x})-\overline{d}(\xi_1,\xi_{k_0}))\right).
\end{eqnarray*}
Then it follows from (\ref{Importance for app}) that
\begin{eqnarray*}
               &&\lambda^{k+1}(x)
     =h_\Xi^k\alpha^k(x)-h_\Xi^k\alpha^k(X_{k_0}^*) \pm \mathcal
               O(n^{-s/d})\\
               &+&
               h_\Xi^k\beta^k(x)\sigma
    \left(w(\overline{d}(\xi_1,{
     x})-\overline{d}(\xi_1,\xi_{k_0}))\right)\\
     &-&
      h_\Xi^k\beta^k(X_{k_0}^*)
     \sigma
    \left(w(\overline{d}(\xi_1,{
     X_{k_0}^*})-\overline{d}(\xi_1,\xi_{k_0}))\right)\\
     &-&
     \left(h_\Xi^k\alpha^k(X_{k_0+1}^*)-h_\Xi^k\alpha^k(X_{k_0}^*)\right.\\
     &+&
     h_\Xi^k\beta^k(X_{k_0+1}^*)
     \sigma
    \left(w(\overline{d}(\xi_1,{
     X_{k_0+1}^*})-\overline{d}(\xi_1,\xi_{k_0}))\right)\\
     &-&
      \left.h_\Xi^k\beta^k(X_{k_0}^*)
     \sigma
    \left(w(\overline{d}(\xi_1,{
     X_{k_0}^*})-\overline{d}(\xi_1,\xi_{k_0}))\right)\right)\\
     &\times&
     \sigma
    \left(w(\overline{d}(\xi_1,{
     x})-\overline{d}(\xi_1,\xi_{k_0}))\right)\\
     &=&
     h_\Xi^{k+1}\alpha^{k+1}(x)\\
     &+&
     h_\Xi^{k+1}\beta^{k+1}(x)\sigma
    \left(w(\overline{d}(\xi_1,{
     x})-\overline{d}(\xi_1,\xi_{k_0}))\right)\\
     &\pm&
     \mathcal
     O(n^{-s/d}),
\end{eqnarray*}
where we use the differentiable property of $\sigma$ and
multivariate mean value theorem in the last equality. Thus,
(\ref{proof approx 3}) holds.  Combining (\ref{proof approx 3}),
(\ref{proof of approx  2}), (\ref{comparison}) with (\ref{app1}) we
obtain for $s> 1$
\begin{equation}\label{Approximation error s>1}
    \mathbf
    E\left[\|\widetilde{N_{n,w}^r}-f_\rho\|_\rho^2\right]\leq
    C'''m^{-2s/(2s+d)},
\end{equation}
where $C'''$ is a constant depending only on $C$, $s$, $\sigma$, $M$
and $r$. Then, Theorem \ref{THEOREM1} follows from (\ref{Error
decomposition}), (\ref{Approximation error s=1}),
(\ref{Approximation error s>1}), (\ref{Sample error r=1}) and
(\ref{sample error r>1}).

\section{Conclusion}\label{Sec. Conclusion}

In this paper, we succeeded in constructing an FNN,  called
\underline{c}onstructive {\underline{f}}eed-forward neural
{\underline{n}}etwork (CFN), for learning purpose. Both theoretical
and
 numerical  results showed that CFN is efficient and effective. The
idea of ``constructive neural networks'' for learning purpose
provided a new springboard for developing scalable neural
network-type learning systems.

We concluded in this paper by  presenting some extensions of the
constructive neural networks learning. In the present paper, the
neural network was constructed by using the method in
\cite{Lin2014b}. Besides \cite{Lin2014b}, there are large portions
of  neural networks  constructed to approximate   smooth functions,
such as
\cite{Anastassiou2011,Cao2008,Costarelli2013,Costarelli2013a}. All
these constructions are proved to possess prominent approximation
capability and simultaneously, suffer from the saturation problem.
We guess that by using the  approach in this paper,
  most of
these neural networks can be used for learning. We will keep
studying in this direction and report the progress
in a future publication.

\section*{Acknowledgement}
The research was supported by the National Natural Science
Foundation of China (Grant Nos. 61502342, 11401462). We are grateful
for Dr. Xiangyu Chang and Dr. Yao Wang for their helpful
suggestions. The corresponding author is Jinshan Zeng.

\end{document}